\documentclass{article}

\usepackage{microtype}
\usepackage{graphicx}
\usepackage{subfigure}
\usepackage{booktabs} 

\usepackage{hyperref}

\usepackage[accepted]{icml2023}

\usepackage{amsmath}
\usepackage{amssymb}
\usepackage{mathtools}
\usepackage{amsthm}
\usepackage{hyperref}
\usepackage{bm}
\usepackage{multirow}

\usepackage[capitalize,noabbrev]{cleveref}

\theoremstyle{plain}
\newtheorem{theorem}{Theorem}[section]
\newtheorem{proposition}[theorem]{Proposition}

\theoremstyle{definition}

\theoremstyle{remark}

\usepackage[textsize=tiny]{todonotes}

\icmltitlerunning{TIPS: \underline{T}opologically \underline{I}mportant \underline{P}ath \underline{S}ampling for Anytime Neural Networks}

\begin{document}

\twocolumn[
\icmltitle{TIPS: \underline{T}opologically \underline{I}mportant \underline{P}ath \underline{S}ampling for Anytime Neural Networks}




\begin{icmlauthorlist}
\icmlauthor{Guihong Li}{ut}
\icmlauthor{Kartikeya Bhardwaj}{qual}
\icmlauthor{Yuedong Yang}{ut}
\icmlauthor{Radu Marculescu}{ut}
\end{icmlauthorlist}

\icmlaffiliation{ut}{Department of ECE, The University of Texas at Austin, Austin, TX}
\icmlaffiliation{qual}{Qualcomm AI Research, San Deigo, CA; {work done while Kartikeya Bhardwaj was at Arm, Inc}}

\icmlcorrespondingauthor{Radu Marculescu}{radum@utexas.edu}

\icmlkeywords{Machine Learning, ICML}

\vskip 0.3in
]



\printAffiliationsAndNotice{}  

\begin{abstract}
Anytime neural networks (AnytimeNNs) are a promising solution to adaptively adjust the model complexity at runtime under various hardware resource constraints. However, the manually-designed AnytimeNNs are biased by designers' prior experience and thus provide sub-optimal solutions. To address the limitations of existing hand-crafted approaches, we first model the training process of AnytimeNNs as a discrete-time Markov chain (DTMC) and use it to identify the paths that contribute the most to the training of AnytimeNNs. Based on this new DTMC-based analysis, we further propose \textit{TIPS}, a framework to automatically design AnytimeNNs under various hardware constraints. Our experimental results show that {TIPS} can improve the convergence rate and test accuracy of AnytimeNNs. Compared to the existing AnytimeNNs approaches, {TIPS} improves the accuracy by 2\%-6.6\% on multiple datasets and achieves SOTA accuracy-FLOPs tradeoffs.
\end{abstract}

\section{Introduction}\label{sec:intro}
In recent years, deep neural networks (DNNs) have been successful in many areas, such as computer vision or natural language processing~\cite{vaswani2017attentiontransformer,dosovitskiy2020imagevit}. However, the intensive computational requirements of existing large models limit their deployment on resource-constrained devices for Internet-of-Things (IoT) and edge applications. To improve the hardware efficiency of DNNs, multiple techniques have been proposed, such as quantization~\cite{qin2020binary,han2015deepquantization}, pruning~\cite{luo2017thinet,han2015learningpruning}, knowledge distillation~\cite{hinton2015distilling}, and neural architecture search (NAS)~\cite{zoph2016neuralnasgrandpa,liu2018dartsnas,stamoulis2019single,li2020edd,li2023zico}.
We note that all these techniques focus on generating \textit{static} neural architectures that can achieve high accuracy under specific hardware constraints.

Recently, anytime neural networks (AnytimeNNs) have been proposed as an orthogonal direction to static neural networks~\cite{huang2017multimsdnet,yu2019autoslim,DBLPBengioBPP15,cvprWangDWYLAG21,yang2022anytime}. AnytimeNNs adjust the model size at runtime by selecting subnetworks from a static supernet~\cite{chen2019you,li2019improved,yu2019autoslim,yu2019universally,yu2018slimmable}. 
Compared to the static techniques, AnytimeNNs can automatically adapt (at runtime) the model complexity based on the available hardware resources. However, the existing AnytimeNNs are manually designed by selecting a few candidate subnetworks. Hence, such hand-crafted AnytimeNNs are likely to miss the subnetworks that can offer better performance. These limitations of existing manual design approaches motivate us to analyze the properties of AnytimeNNs and then provide a new algorithmic solution. 
Specifically, in this work, we address two \textbf{key questions:}
\begin{enumerate}
    \item \textit{How can we quantify the importance of various operations (\textit{e.g.}, convolutions, residual additions, etc.) to the convergence rate and accuracy of AnytimeNNs?}
    \item \textit{Are there topological (\textit{i.e.}, related to network structure) properties that can help us design better AnytimeNNs?}
\end{enumerate}

\begin{figure}[tb]
\vspace{0mm}
    \centering
    \includegraphics[width=0.36\textwidth]{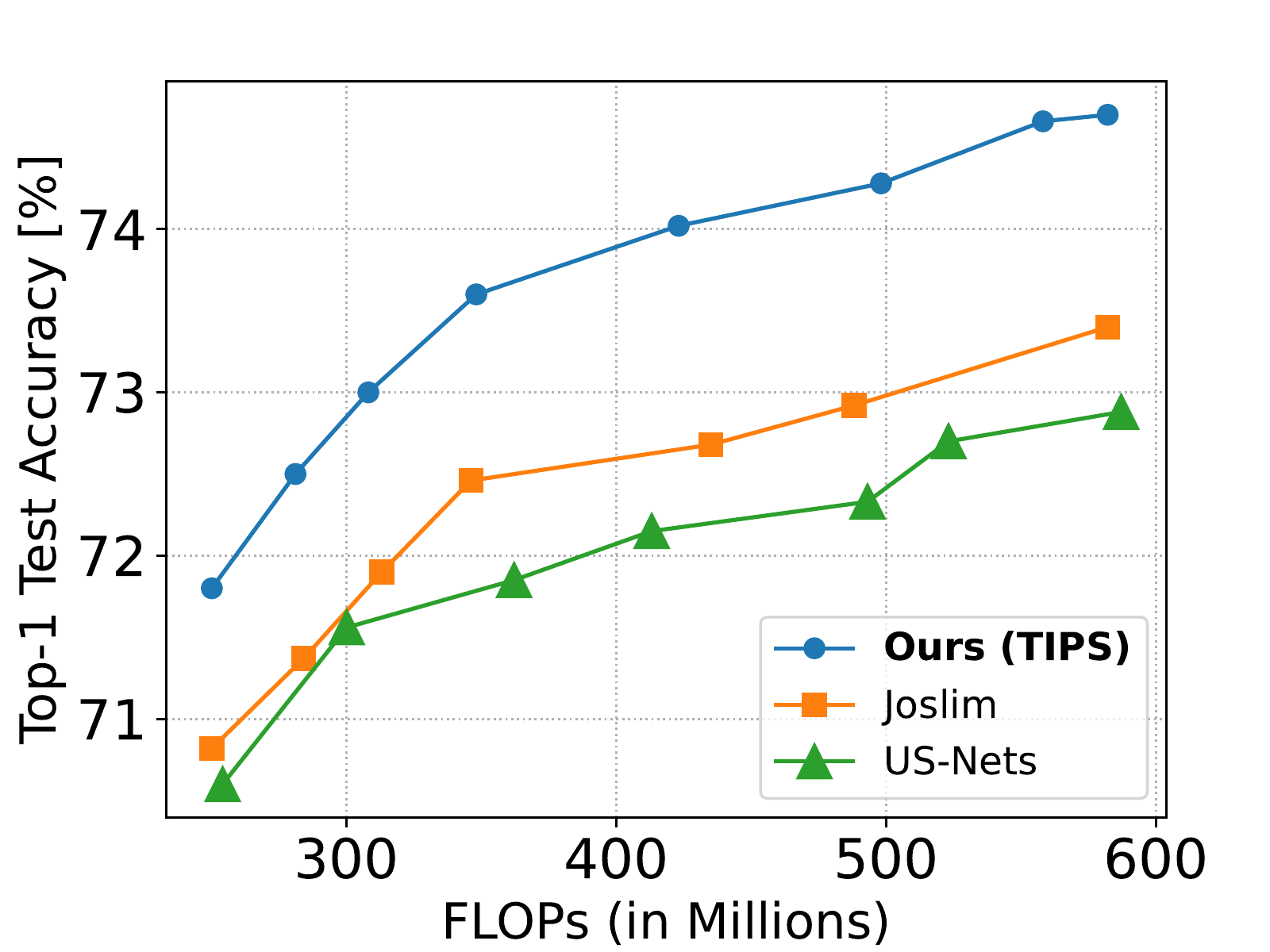}\vspace{-2mm}
    \caption{Test Accuracy vs. FLOPs on ImageNet. {TIPS} achieves higher accuracy (given the same or even fewer FLOPs) than SOTA AnytimeNNs: Joslim~\cite{chin2021joslimrudydiana} and US-Nets~\cite{yu2019universally}.\vspace{-5mm}}
    \label{fig:first}
\end{figure}

To answer these questions, we analyze the AnytimeNNs from a \textit{graph theory} perspective. This idea is motivated by the observation that the topological features of DNNs can accurately indicate their gradient propagation properties and test performance~\cite{bhardwaj2021doescvprnnmass, li2021flash}. 
Inspired by the network structure analysis, given an AnytimeNN, we propose a Discrete-Time Markov Chain (DTMC)-based framework to explore the relationships among different subnetworks. We then propose two new topological metrics, namely \textit{Topological Accumulated Score} (TAS) and \textit{Topological Path Score} (TPS) to analyze the gradient properties of AnytimeNNs. Based on these two metrics, we finally propose a new training method, \textit{i.e.}, \textit{Topologically Important Path Sampling} (TIPS), to improve the convergence rate and test performance of AnytimeNNs. The experimental results show that our proposed approach outperforms SOTA approaches by a significant margin across many models and datasets (see Fig.~\ref{fig:first}). Overall, we make the following \textbf{key contributions:}
\begin{itemize}\vspace{-3mm}
    \item We propose a new importance analysis framework by modeling the AnytimeNNs as DTMCs; this enables us to capture the relationships among different subnetworks of AnytimeNNs. \vspace{-1mm}
    \item Based on the DTMC-based framework, we propose two new topological metrics, \textit{Topological Accumulated Score} (TAS) and \textit{Topological Path Score} (TPS), which can characterize the operations that contribute the most to the training of AnytimeNNs. \vspace{-1mm}
    \item We propose a new theoretically-grounded training strategy for AnytimeNNs, namely, \textit{Topologically Important Path Sampling} ({TIPS}), based on our importance analysis framework. We show that {TIPS} achieves a faster convergence rate compared to SOTA training methods.  \vspace{-3mm}
    \item We demonstrate that {TIPS} enables the automatic design of better AnytimeNNs under various hardware constraints. Compared to existing AnytimeNN methods, TIPS improves the accuracy by 2\%-6.6\% and achieves SOTA accuracy-FLOPs tradeoffs on multiple datasets, under various hardware constraints (see Fig.~\ref{fig:first}). \vspace{-1mm}
\end{itemize}\vspace{-2mm}
The rest of the paper is organized as follows. In Section~\ref{sec:related}, we discuss related work. 
In Section~\ref{sec:approach}, we formulate the problem and introduce our proposed solution ({TIPS}). 
Section~\ref{sec:experiment} presents our experimental results and outline directions for future work. Finally, Section \ref{sec:conclusion} concludes the paper.

\section{Related Work}\label{sec:related}\vspace{-2mm}
There are three major directions related to our work:
\subsection{Anytime Inference}\vspace{-1mm}
Anytime neural networks (AnytimeNNs) can adapt the model complexity at runtime to various hardware constraints; this is achieved by selecting the optimal subnetworks of a given (static) architecture (\textit{supernet}), while maintaining the test accuracy. The runtime adaptation of AnytimeNNs is primarily driven by the available hardware resources~\cite{YuanWSLZB20s2dnas}. 
For instance, early-exit networks can stop the computation at some intermediate layers of the supernet and then use individual output layers to get the final results~\cite{wang2018skipnet,veit2018convolutionalskipnet2, rebuttal_exit1}. 
Similarly, skippable networks can bypass several layers at runtime~\cite{ieee_rice,larsson2016fractalnetdepth1,rebuttal_depth1}. 
Alternatively, approaches for slimmable networks remove several channels of some layers at runtime~\cite{lee2018anytimewidth1,Bejnordi2020Batchshaping,yang2018netadapt,hua2018channel,li2021dynamic_dsnet,chin2021joslimrudydiana,rebuttal_width1,rebuttal_width2}. Finally, multi-branch networks select the suitable branches of networks to reduce the computation workload to fit the current hardware constraints~\cite{cai2021dynamic,aaaiRuizV21,huang2017multimsdnet,rebuttal_branch1}.

\subsection{Layerwise Dynamical Isometry (LDI)}\vspace{-2mm}
LDI is meant to quantify the gradient flow properties of DNNs~\cite{saxe2013exact,xiao2018dynamicalldi,burkholz2019initializationldi}. For a deep neural network, let ${x_i}$ be the output of layer $i$; the Jacobian matrix of layer $i$ is defined as: 
$J_{i,i-1} = \frac{\partial {x_{i}}}{\partial {x_{i-1}}}$.
Authors of~\cite{Lee2020Aldi} show that if the singular values of $J_{i,i-1}$ for all $i$ at initialization are close to 1, then the network satisfies the LDI, and the magnitude of the gradient does not vanish or explode, thus benefiting the training process. 

\subsection{Network Topology} Previous works show that the topological properties can significantly impact the convergence rate and test performance of deep networks. For example, by modeling deep networks as graphs, authors in ~\cite{bhardwaj2021doescvprnnmass} prove that the average node degrees of deep networks are highly correlated with their convergence speeds. Lately, \cite{chen2022deep} developed a similar understanding of neural networks' connectivity patterns on its trainability. Moreover, several works also show that some specific topological properties of deep networks can indicate their test accuracy~\cite{li2021flash,javaheripi2021swann}. We note that these existing approaches primarily focus on networks with a static structure. The relationship between topological properties and the convergence/accuracy of networks with varying architectures (\textit{e.g.}, AnytimeNNs) remains an open question. This motivates us to investigate the topological properties of AnytimeNNs.

\begin{figure*}[t!]
    \centering
    \includegraphics[width=0.98\textwidth]{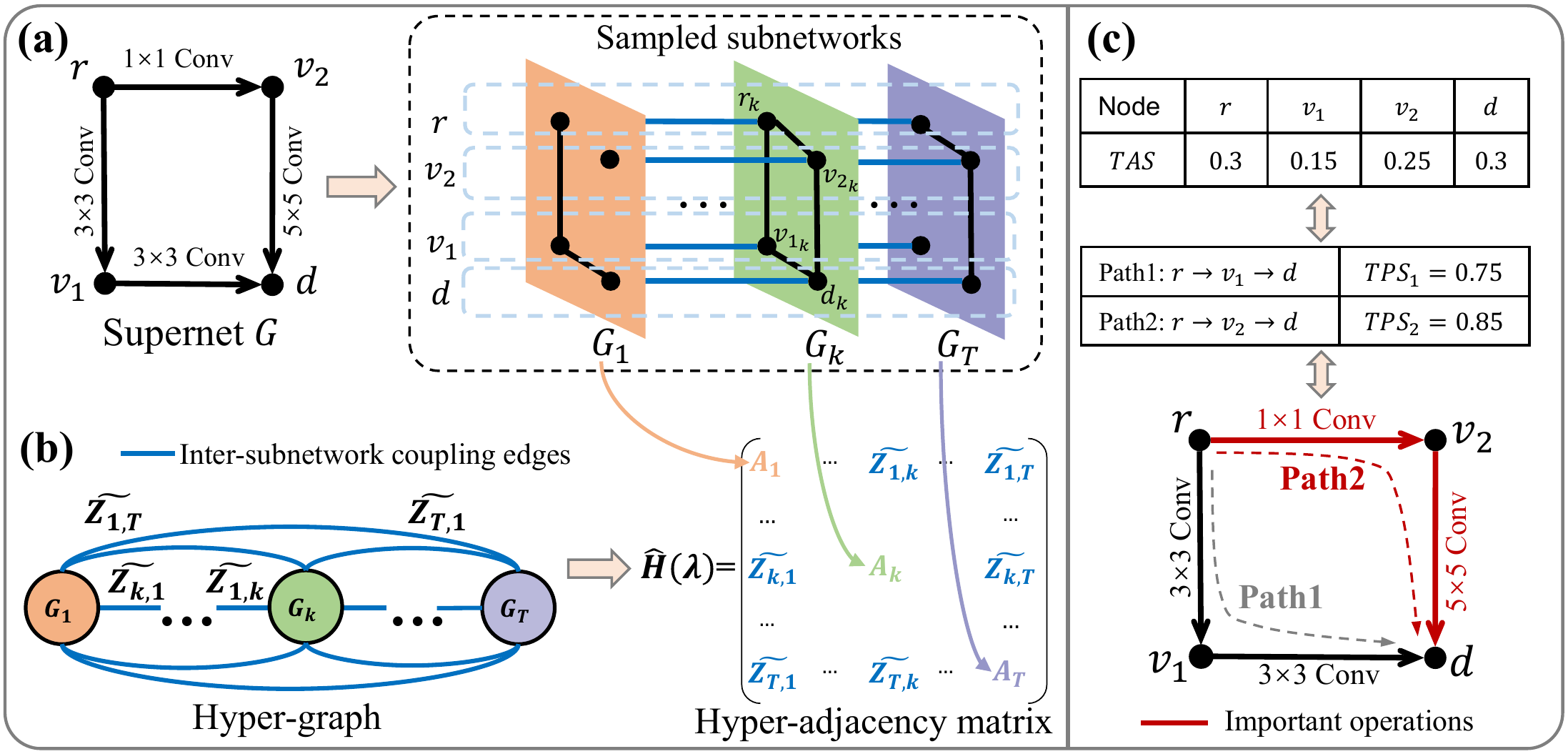}\vspace{-2mm}
    \caption{Overview of our proposed DTMC-based analysis (\textbf{a}) We model the operations (\textit{e.g.}, 1$\times$1-Conv) as edges; we model the outputs of such operations (\textit{i.e.}, featuremaps) as nodes (\textit{e.g.}, $r,v_2$); here $r$ and $d$ are the input and output nodes of the supernet $G$, respectively. By removing some edges from supernet $G$, we generate multiple subnetworks \{$G_k,\ k=1,...,T$\}. (\textbf{b}) We then build the adjacency matrices \{$\bm{A}_k,\ k=1,...,T$\} for each subnetwork. We combine these adjacency matrices and the inter-subnetwork coupling matrices \{$\widetilde{\bm{Z}_{i,j}},\ i\neq j$\} to form the hyper-adjacency matrix $\hat{\bm{H}}(\lambda)$. (\textbf{c}) By solving Eq.~(\ref{eq:normalmatrix})(\ref{eq:properyofdtmc})(\ref{eq:deftas}), we get the TAS value of each node. After finding the path with the highest TPS value by Eq.~\ref{eq:def_tps} (Path2), we characterize the important operations (\textit{i.e.}, edges) of the AnytimeNN. We provide more examples in Appendix~\ref{app:illustrate}.\vspace{-4mm}}
    \label{fig:overview}
\end{figure*}

\section{Approach}\label{sec:approach}
In this work, for a given deep network (\textit{i.e.}, supernet), our goal is to automatically find its AnytimeNN version under various hardware constraints. To this end, our approach consists of three major steps: (\textit{i}) Characterize the importance of each operation (convolution, residual addition, \textit{etc.}). As such, we model the training process of AnytimeNNs as a DTMC and use it to analyze their topological properties (Fig.~\ref{fig:overview}). (\textit{ii}) Based on this importance analysis, we then propose a new training strategy ({TIPS}) to improve the accuracy of AnytimeNNs. (\textit{iii}) Finally, we search for the AnytimeNNs under various hardware constraints. Next, we discuss these steps in detail.

\subsection{Modeling AnytimeNNs as Markov Chains}\label{sec:approach_model}
\subsubsection{Modeling AnytimeNNs as Graphs}
As shown in Fig.~\ref{fig:overview}(a), we model various DNN operations (convolutions, residual additions, \textit{etc.}) as \textit{edges}, and model the outputs of such operations (\textit{i.e.}, featuremaps) as \textit{nodes} in a graph. This way, a given architecture (supernet) is represented by a static undirected graph
$G = (V,E)$, where $V$ is the set of nodes (with $|V | = N$) and $E$ is the set of edges between nodes (with $|E| = M$). 
For a given DNN architecture, its corresponding AnytimeNNs select suitable subnetworks under the current hardware constraints. Specifically, these \textit{subnetworks} $G_k$ are obtained by sampling edges from the initial supernet $G$:
\begin{equation}
    G_k = (V, E_k), E_k\subseteq   E
\vspace{0mm}
\end{equation}
where, the node set $V$ is the same for all subnetworks, but different subnetworks $G_k$ have different edge sets $E_k$ (see Fig.~\ref{fig:overview}(a)). 
To ensure that the sampled subnetwork is valid, we always sample the input, output, and down-sample layers (\textit{e.g.}, layers with pooling or stride=2). To ensure the validity of a subnetwork, we first randomly decide whether or not each layer remains in the subnetwork. For the remaining layers, we also ensure that \#channels in consecutive layers match. As shown in Fig.~\ref{fig:overview}(a), based on the topology of $G_k$, we can construct the adjacency matrix $\bm{A}_k\in R^{N \times N }$ for a subnetwork as follows:
\begin{equation}\label{eq:adj_mat_cases}
    \left\{
\begin{aligned}
\bm{A}_{k}(s,t)=0,&&\text{if}\ (s,t)\notin E_k \\
\bm{A}_{k}(s,t)=1,&&\text{if}\ (s,t)\in E_k\\
\bm{A}_{k}(s,t)=1,&&\text{if}\ s=t=1\ \text{or } s=t=N\\
\end{aligned}
\right.
\end{equation}
where each edge $(s,t)$ corresponds to an operation in the given network. The intuition behind the values of $\bm{A}_{k}(1,1)$ and $\bm{A}_{k}(N,N)$ is that the computation always starts from the input/output layer in the forward/backward path. We note that our objective is to analyze the \textit{layer-wise} gradient properties of AnytimeNNs. Since the singular values of each layer's Jacobian are designed to be around 1 by commonly used initialization schemes (\textit{e.g.}, by maintaining uniform gradient variance at all layers), it is reasonable to assign `1' as the weight of each edge (\textit{i.e.}, operation) in Eq.~\ref{eq:adj_mat_cases} if it appears in $A_k$. More details are given in Appendix~\ref{app:illustrate}.

At each training step, we sample $T$ subnetworks as shown in Fig.~\ref{fig:overview}(a). Let $\mathcal{L}$ denote the loss function (\textit{e.g.}, cross-entropy). Then, the loss for AnytimeNNs at each training step is calculated by passing the same batch of images through these $T$ subnetworks~\cite{huang2017multimsdnet,mutilelosstrain}:
\begin{equation}\label{eq:acc_loss}
\vspace{0mm}
    Loss = \sum_{k=1}^T \mathcal{L}(y,G_k(x))    
\vspace{0mm}
\end{equation}
where $x$, $y$, $G_k(x)$ are the input, ground truth, and output of subnetwork $G_k$, respectively.
Eq.~\ref{eq:acc_loss} shows that all these subnetworks in Fig.~\ref{fig:overview}(a) share the same input data and use the accumulated loss from all of them to calculate the gradient during the backward propagation.
Hence, all these subnetworks are highly coupled with each other.  
Inspired by the idea in~\cite{taylor2019supracentrality}, as shown in Fig.~\ref{fig:overview}(b), we integrate multiple subnetworks into a new hyper-graph to capture the coupling impacts among different subnetworks. Specifically, given a sequence of $T$ subnetworks and each subnetwork with $N$ nodes, we construct a \textit{hyper-adjacency matrix} $\hat{\bm{H}}(\lambda)\in R^{NT \times NT}$:
\begin{equation}\label{eq:superadj}
    \hat{\bm{H}}(\lambda)= \begin{bmatrix}
{\bm{A}_1} &\widetilde{\bm{Z}_{1,2}}&...  & \widetilde{\bm{Z}_{1,T}}\\ 
 \widetilde{\bm{Z}_{2,1}}&{\bm{A}_2}&...  &\widetilde{\bm{Z}_{2,T}} \\ 
...&...&...&...\\
 \widetilde{\bm{Z}_{T,1}}&...& ... & {\bm{A}_T}
\end{bmatrix}
\vspace{0mm}
\end{equation}
where $\widetilde{\bm{Z}_{i,j}}\in R^{N\times N}$ is the inter-subnetwork coupling matrix between \textit{different} subnetworks $G_i$ and $G_j$ as follows:
\vspace{0mm}
\begin{equation}\label{eq:InterLayer}
    \widetilde{\bm{Z}_{i,j}}=\lambda\bm{I}, \ i\neq j\ , \ 0<\lambda\leq1
\vspace{0mm}
\end{equation}
where $\bm{I}$ is the identity matrix. 

\textit{Remark}: On the one hand, $\bm{A}_k$ in $\hat{\bm{H}}(\lambda)$ capture the connectivity pattern of each individual subnetwork. On the other hand, as shown in Fig.~\ref{fig:overview}(a)(b), $\widetilde{\bm{Z}_{i,j}}$ in $\hat{\bm{H}}(\lambda)$ captures the inter-subnetwork coupling effects between every pair of subnetworks by connecting same nodes across different subnetworks\footnote{The parameter $\lambda$ controls the strength of the interactions between different subnetworks; see details in Sec~\ref{sec:exp_ablation}.}. Hence, our methodology does capture both \textit{intra-} and \textit{inter-subnetwork} topological properties. This is crucial since AnytimeNNs have a variable network architecture. (see more discussion in Section~\ref{sec:exp_ablation} and Appendix~\ref{app:detail_tips}).

\subsubsection{Building the DTMC for AnytimeNNs}
In this work, we aim to identify the importance of each operation in AnytimeNNs. Inspired by the PageRank algorithm~\cite{berkhin2005surveypagerank}, we use the \textit{hyper-adjacency matrix} $\hat{\bm{H}}(\lambda)$ to build the transition matrix $\bm{P}$ of our DTMC. Specifically, we normalize the adjacency matrix $\hat{\bm{H}}$ row by row: 
\begin{equation}\label{eq:normalmatrix}
\bm{P}_{m,:}=\hat{\bm{H}}_{m,:}(\lambda)/({\sum}{_n}{\hat{\bm{H}}_{m,n}(\lambda)})
\end{equation}
and obtain an irreducible, aperiodic, and homogeneous DTMC, which has a unique stationary state distribution ($\bm{\pi}$)~\cite{hajek2015randombookmarkov}\footnote{More details are given in Appendix~\ref{app:dtmc}}. The stationary distribution $\bm{\pi}$ of such DTMC has the following property:
\begin{equation}\label{eq:properyofdtmc}
\bm{\pi} \bm{P}=\bm{\pi}
\end{equation}
Hence, we can solve Eq.~\ref{eq:properyofdtmc} to obtain $\bm{\pi}$ for our DTMC.  Next, we use $\bm{\pi}$ to analyze the nodes in $\hat{\bm{H}}(\lambda)$.  
We denote $\bm{\pi}(s)$ as the stationary probability of a state $s$. 
Note that, as shown in Fig.~\ref{fig:overview}(a), a node $r$  appears in all the $T$ sampled subnetworks, hence it appears $T$ times in $\hat{\bm{H}}(\lambda)$; each node $r$ from the supernet $G$ corresponds to $T$ nodes $\{r_k,\ k=1,...,T\}$ in the DTMC within $T$ subnetworks $\{G_k,\ k=1,...,T\}$. 
For a given node $r_k$ in the DTMC, we denote its stationary probability as $\bm{\pi}(s_{r_k})$. 

\subsection{Topological \& Gradient Properties of AnytimeNNs}\label{sec:approach_nnpath}
To analyze the importance of nodes and paths in AnytimeNNs, we propose the following definitions:

\textbf{Definition 1. Topological Accumulated Score (TAS)} A \textit{topological accumulated score} of a node $r$ from the supernet is its accumulated PageRank score across multiple subnetworks. For a given node $r$ in $V$, its TAS value $\mu_r$ is:
\begin{equation}\label{eq:deftas}
    \mu_r = \sum_{k=1}^{T}\bm{\pi}(s_{r_k})
\vspace{0mm}
\end{equation}
TAS quantifies the \textit{accumulated probability} that a node is selected within an AnytimeNN. Next, we use TAS to analyze the importance of various computation paths.

\textbf{Definition 2. Topological Path Score (TPS)} In an AnytimeNN, we define a computation path $l$ from a node $r$ to a node $d$, as a sequence of edges $\{r \rightarrow v_1 \rightarrow  ...  \rightarrow v_w \rightarrow d\}$. The \textit{topological path score} $TPS_l$ of a computation path $l$ is the sum of the TAS values of \textit{all} nodes traversed in the path:
\begin{equation}\label{eq:def_tps}
   TPS_l=\sum_{s\in\{r,v_1,...,v_w,d\}}\mu_{s}
\vspace{0mm}
\end{equation}

The above definitions and the LDI discussion in Section~\ref{sec:related} enable us to propose our main result:

\begin{proposition}\label{prop:dynamic_networks}
Consider an AnytimeNN initialized by a zero-mean i.i.d. distribution with variance $q$. Given two computation paths $l_S$ and $l_L$ in this AnytimeNN with same width $w_r$ and number of nodes $D$, we define $w_e^S$ ($w_e^L$) as the average degree of $l_S$ ($l_L$). Assuming $q\leq\epsilon$, $w_e^S\gg w_r$, and $w_e^L\gg w_r$, then the mean singular values $E[\sigma^S]$ and $E[\sigma^L]$ of the Jacobian matrix for $l_S$ and $l_L$ satisfy:
\begin{equation}
    \text{if }\ TPS_{l_S}\leq TPS_{l_L}\text{, then } E[\sigma ^S]\leq E[\sigma^L]\leq 1
\end{equation}
where, $\epsilon= \frac{1}{max(w_e^S,w_e^L)+w_r + 2\sqrt{max(w_e^S,w_e^L)w_r}}$. That is, the mean singular value of the Jacobian for the computation path with higher TPS values is higher and closer to 1.
\end{proposition}

\begin{figure}[t!]
    \centering
    \includegraphics[width=0.45\textwidth]{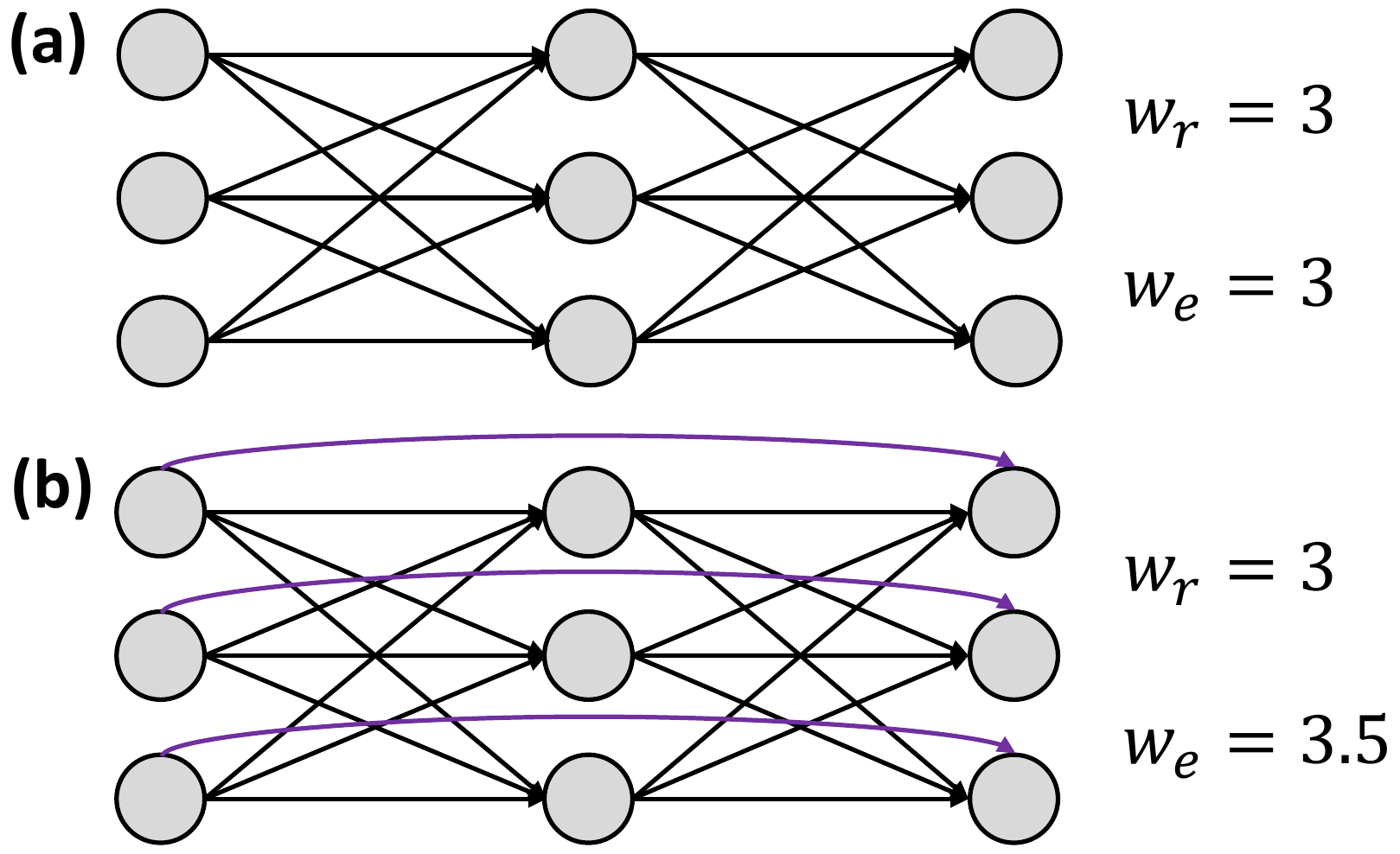}
    \caption{A 2-layer MLP has three neurons per layer; the right version includes skip connections (purple) across layers, while the left does not. Both MLPs have a real width $w_r$ of 3. The average degree $w_e$ is calculated as the total links (weights and skip connections) divided by total neurons (excluding input neurons). The upper MLP in \textbf{(a)} has a $w_e$ of $18/6=3$, and the lower in \textbf{(b)} has a $w_e$ of $21/6=3.5$ due to skip connections.}
    \label{fig:w_e_wr}
\end{figure}

\textit{Proof.}
Authors in~\cite{bhardwaj2021doescvprnnmass} prove that for a given neural network initialized by a zero-mean i.i.d. distribution with variance $q$, the mean singular value $E[\sigma]$ of the Jacobian matrix from the network is bounded by the following inequality:
\begin{equation}\label{eq:ldi_bound}
    \sqrt{qw_e} -\sqrt{qw_r} \leq E[\sigma] \leq \sqrt{qw_e}+\sqrt{qw_r}
\end{equation}
where the $w_e$ is the \textit{average node degree} or \textit{effective width} and $w_r$ is the real width of the neural network. In Fig.~\ref{fig:w_e_wr}, we give an example of how $w_e$ and $w_r$ are calculated in a neural network.

We now use the above bounds to prove our main result. Let us first prove the right side of the inequality in Proposition~\ref{prop:dynamic_networks}. According to~Eq.~\ref{eq:ldi_bound}, for two computational paths $l_S$ and $l_L$, the mean singular values $E[\sigma^S]$ and $E[\sigma^L]$ of their Jacobian matrices are bounded by the following inequalities:
\begin{equation}\label{eq:bi_small_ldi_bound}
\begin{aligned}
        \sqrt{qw_e^S} -\sqrt{qw_r} \leq E[\sigma^S] \leq \sqrt{qw_e^S}+\sqrt{qw_r}\\
        \sqrt{qw_e^L} -\sqrt{qw_r} \leq E[\sigma^L] \leq \sqrt{qw_e^L}+\sqrt{qw_r}
\end{aligned}
\end{equation}

Based on Eq.~\ref{eq:bi_small_ldi_bound}, we note that if initialization variance $q$ satisfies
\begin{equation}\label{eq:var_bound} 
q\leq\frac{1}{max(w_e^L,w_e^S)+w_r + 2\sqrt{max(w_e^L,w_e^S)w_r}}  
\end{equation}
then, the mean singular value is always bounded by 1 for both $l_S$ and $l_L$; that is:
\begin{equation}\label{eq:ldi1}
    E[\sigma^S]\leq1 \ \  \ \  \ \  E[\sigma^L]\leq1
\end{equation}

Inequality~\ref{eq:ldi1} proves the right side of inequality in Proposition~\ref{prop:dynamic_networks}. Next, we prove the left side.

Using Eq.~\ref{eq:bi_small_ldi_bound}, if $w_e^S\gg w_r$ and $w_e^L\gg w_r$, then the mean singular values of $l_L$ and $l_S$ are mainly determined by $w_e^L$ and $w_e^S$; that is: 
\begin{equation}\label{eq:approxldi}
    E[\sigma^L]=\sqrt{qw_e^L}\ \  \ \  \ \   E[\sigma^S]=\sqrt{qw_e^S}
\end{equation}

From Definition~1, we know that TAS of a given node is the sum of its PageRank across the $T$ subnetworks. As discussed in~\cite{fortunato2006approximatingpagerank}, under the mean-field approximation, the PageRank of a given node is linearly correlated to its node degree. That is, for the $i^{th}$ node $i$ on the computation path:
\begin{equation}\label{eq:indeg_rank}
    \mu_i = \frac{k_i}{C}
\end{equation}
where $k_i$ is the node degree for node $i$, and $C$ is a constant determined by the topology of supernet\footnote{A supernet is the given neural architecture that needs to be converted into its AnytimeNN version.}. Because both $l_S$ and $l_L$ are sampled from the same supernet, then they share the same value of $C$.
Combining Eq.~\ref{eq:indeg_rank} with Definition~2, the TPS satisfies the following relation:
\begin{equation}\label{eq:tps_ln_raw}
    TPS= \sum_{i=1}^D \mu_i = \frac{\sum_{i=1}^D k_i}{C}
\end{equation}
Given the definition of average degree $w_e$, we rewrite Eq.~\ref{eq:tps_ln_raw} as follows:
\begin{equation}\label{eq:tps_ln_inter}
    TPS=  \frac{D\times w_e}{C} \Longrightarrow w_e = \frac{C\times TPS}{D}
\end{equation}
where $D$ is the number of nodes for a given path.
By combining Eq.~\ref{eq:approxldi} and Eq.~\ref{eq:tps_ln_inter}, the mean singular value is determined by $q$, $C$, $D$, and $TPS$:
\begin{equation}\label{eq:tps_ln}
    w_e = \frac{C\times TPS}{D}\Longrightarrow E[\sigma] = \sqrt{\frac{q\times C\times TPS}{D}}
\end{equation}
Note that $q$, $C$, $D$ have the same values for both $f_S$ and $f_L$. Hence, if $TPS_S\leq TPS_L$, then $E[\sigma^S]\leq E[\sigma^L]$. This proves the left side of inequality in Proposition~\ref{prop:dynamic_networks}.

Therefore, the inequality in Proposition~\ref{prop:dynamic_networks} holds true for both the left and right sides. That is, the mean singular value of the Jacobian for a computation path with a higher TPS is higher and closer to 1. Moreover, the closeness of $E[\sigma]$ to 1 is determined by the initialization variance $q$, constant $C$, the values of $TPS_S$ and $TPS_L$, and \#nodes $D$. This completes our proof of Proposition~\ref{prop:dynamic_networks}. \qed

Intuitively, Proposition~\ref{prop:dynamic_networks} says that the computation paths with high TPS values satisfy the LDI property and the gradient magnitude through such paths would not vanish or explode, thus, having a higher impact on AnytimeNNs training. We provide empirical results to verify this in the experiments section.

\vspace{-0.5mm}\begin{algorithm}[thb]
 {\small
  \caption{Pareto-optimal subnetwork search}
   \label{alg:paretosearch}
\begin{algorithmic}
\STATE {\bfseries Input:} Supernet $G$, search steps $m$
\STATE {\bfseries Output:} Pareto-optimal subnetworks set $G_P$
\STATE {\bfseries Search:}

\STATE Initialize $G_P = \phi$
\FOR{$i=1$ {\bfseries to} $m$}
\STATE Sample subnetwork $G_i$ from $G$
\STATE Evaluate $G_i$ and get its accuracy $\Theta_{G_i}$ 
\STATE ${optimal=TRUE}$
\STATE Initialize false-Pareto Set $G_{P_{out}}=\phi$
\FOR{ $G_j$ in $G_P$}
\IF{${FLOPs}_{G_j}\leq {FLOPs}_{G_i}$ \AND $\Theta_{G_j}>\Theta_{G_i}$}
\STATE ${optimal=FALSE}$
\ELSIF{${FLOPs}_{G_j}\leq {FLOPs}_{G_i}$ \AND $\Theta_{G_j}<\Theta_{G_i}$}
\STATE Add $G_j$ to $G_{P_{out}}$
\ENDIF
\ENDFOR
\IF{${optimal}$}
\STATE Add $G_i$ to $G_P$
\ENDIF
\STATE Remove false Pareto-optimal $G_P=G_P\setminus G_{P_{out}}$
\ENDFOR
\end{algorithmic}}
\end{algorithm}
\vspace{-3mm}

\subsection{Topologically Important Path Sampling (TIPS)}\label{sec:TIPS}\vspace{-2mm}
Among the computation paths with the same number of nodes of an AnytimeNN, we define the operations (\textit{i.e.}, edges) along the path with the highest TPS value as \textit{important operations}; the rest of operations are deemed as \textit{unimportant operations} (see Path2 in Fig.~\ref{fig:overview}(c)). According to Proposition~\ref{prop:dynamic_networks}, the path with higher TPS values has higher singular values of the Jacobian matrix. Hence, these {important operations} have a significant impact on the training process. Note that, previous works treat all operations uniformly~\cite{chin2021joslimrudydiana,wang2018skipnet}. Instead, in our approach, we modify the sampling process during the training process and use a higher sampling probability to sample these important operations. We call this sampling strategy \textbf{\textit{Topologically Important Path Sampling}} (\textbf{\textit{TIPS}}). More details are given in the experiments section.

\subsection{Pareto-Optimal Subnetwork Search}\label{sec:apporach_pareto_search}\vspace{-2mm}
After the {TIPS}-based training, we use the Algorithm~\ref{alg:paretosearch} to search for the Pareto-optimal subnetworks under various hardware constraints. To this end, we consider the number of floating-point operations (FLOPs) as a proxy for hardware resource constraints\footnote{In practice, this can be easily replaced by some other hardware resource, such as memory or power consumption.}. At runtime, one can select the proper subnetworks to quickly adapt to various hardware constraints; \textit{e.g.}, if the amount of currently available memory for a device drops below a threshold, we switch to a smaller subnetwork to meet the new memory budget.

\subsection{Summary of Our Approach} 
In brief, our method consists of the following steps: 
\vspace{-3mm}
\begin{itemize}
\item \textbf{Step 1: TPS analysis} (Fig.~\ref{fig:overview}) We sample subnetworks and exploit TPS (our DTMC-based metric for AnytimeNNs) to identify important operations. \vspace{-3mm}
\item \textbf{Step 2: AnytimeNN training} We use {TIPS} by assigning a higher sampling probability to the important operations (as given by TPS) to \textit{train} AnytimeNNs. \vspace{-3mm}
\item \textbf{Step 3: Pareto-optimal search} Before model deployment, we do an \textit{offline} search under \textit{various} hardware constraints. We store the full supernet and \#channel configurations of the obtained subnetworks.\vspace{-3mm}
\end{itemize}
\vspace{-0mm}

\textbf{\textit{Remarks}}: Our framework involves two steps where we perform sampling. In Step 1, we conduct the TAS and TPS analysis without knowing the importance of each operation. Hence, to build the DTMC with the sampled subnetworks (Section 3.1), we uniformly sample these operations and ensure each operation is selected at least once during this stage. Once we compute the TAS and TPS values, we can identify the important and unimportant operations in a given supernet (Sections 3.2 and 3.3). During Step 2 (i.e., AnytimeNN training), operations are not sampled uniformly. Instead, important operations are sampled with a higher probability compared to unimportant operations. We provide more details in Section~\ref{sec:exp_val_tips}.

After Steps 1-3, \textit{at runtime}, we use the best subnetwork configurations under various budgets. We provide the storage overhead and time efficiency analysis in Appendix~\ref{app:overhead}.

In terms of \textbf{time cost}: Step 1 takes only 39 seconds on a Xeon CPU for MobileNet-v2. Step 2 takes 97 hours on an 8-GPU server for 150 epochs, and Step 3 takes 8 minutes on an RTX3090 GPU. 
We note that Step 1 only has negligible time costs compared to AnytimeNN training. 
Moreover, Steps 1-3 are conducted offline and, hence, they result in zero overhead for the online inference.

\section{Experimental Results}\label{sec:experiment}

\begin{figure*}[htb]
\centering

  \subfigure[EfficientNet-B0]{\includegraphics[width=0.3\textwidth]{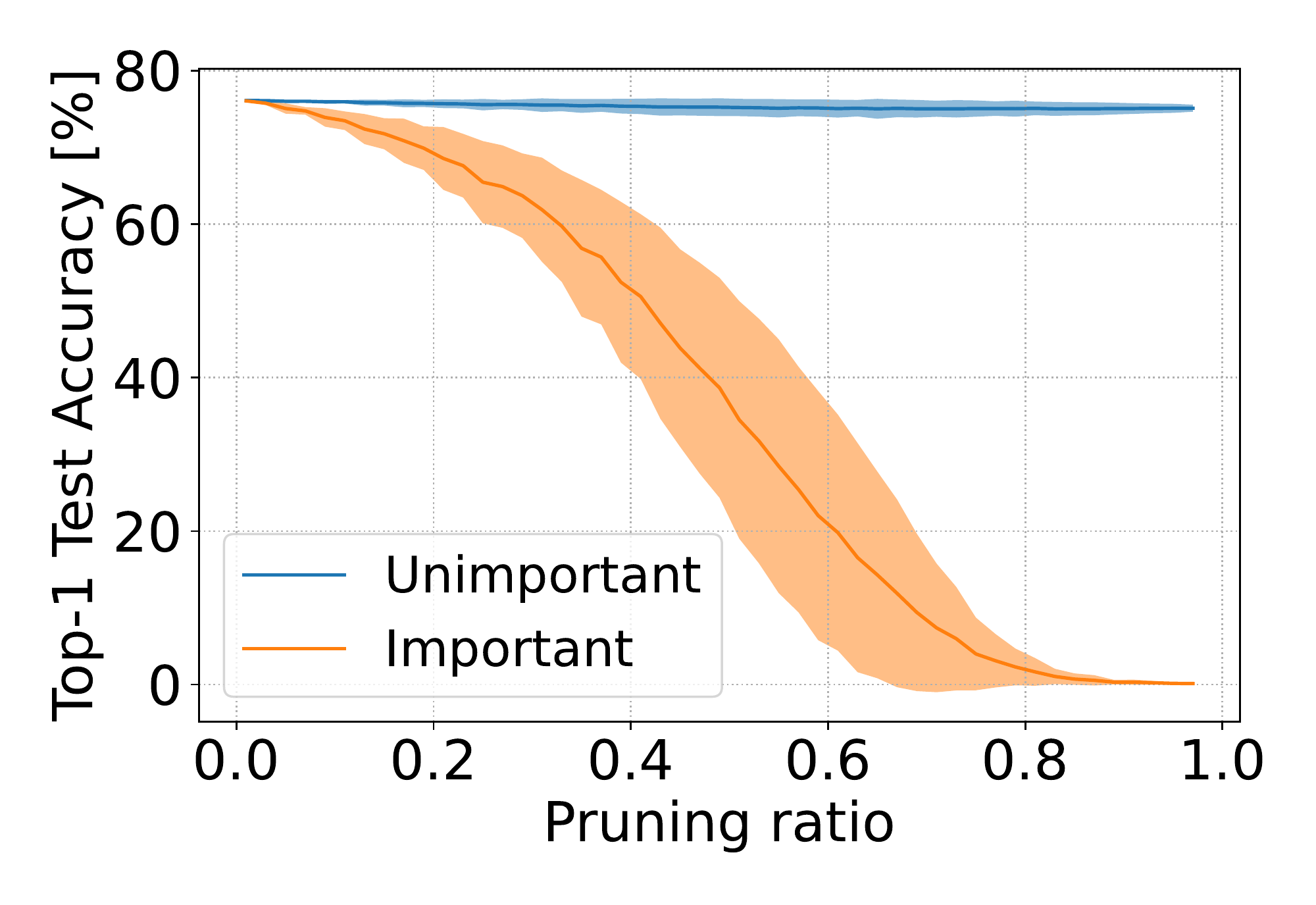}\vspace{-4mm}}
   \hfill	
  \subfigure[ResNet18]{\includegraphics[width=0.3\textwidth]{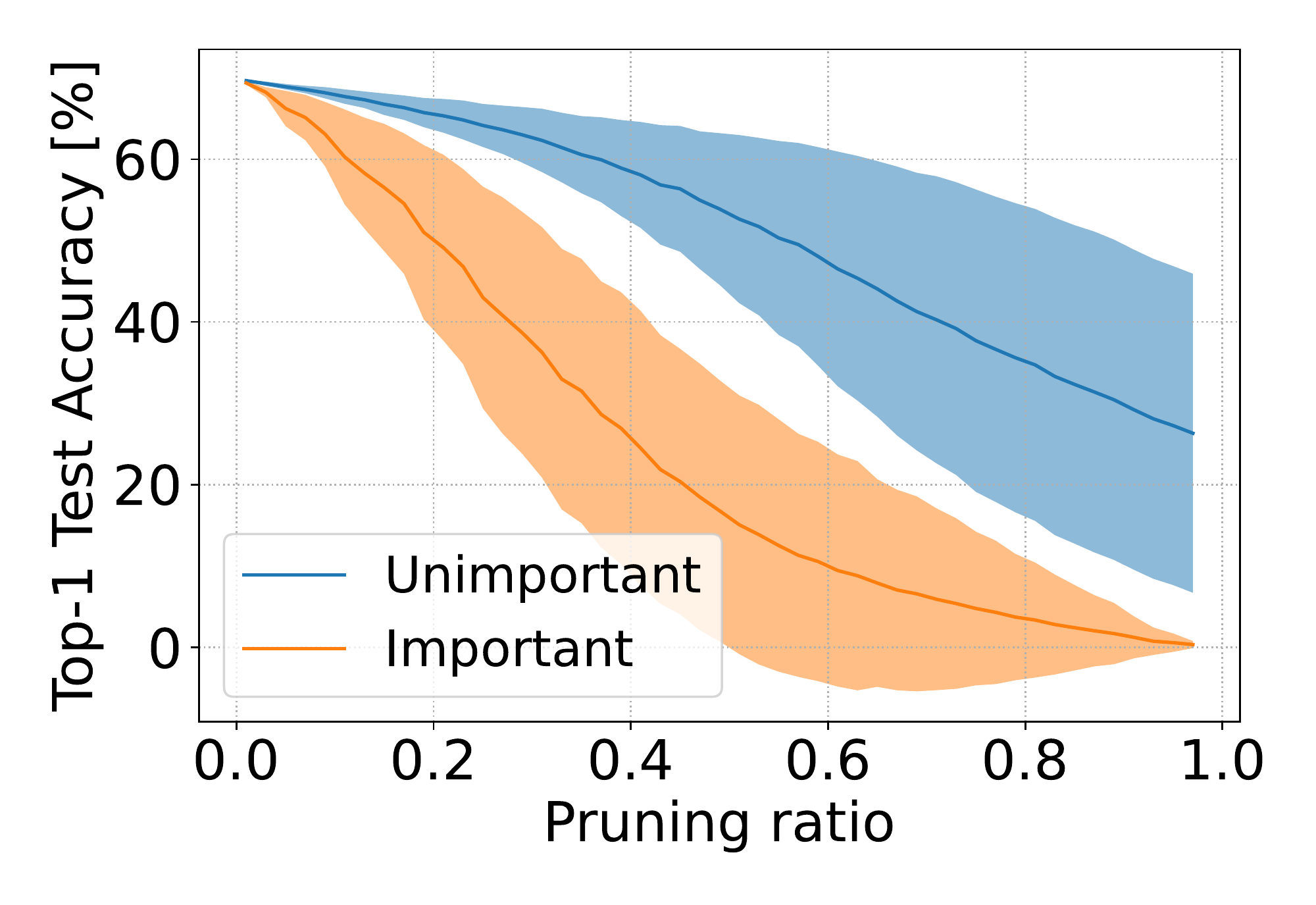}\vspace{-4mm}}
   \hfill	
  \subfigure[MobileNet-v2]{\includegraphics[width=0.3\textwidth]{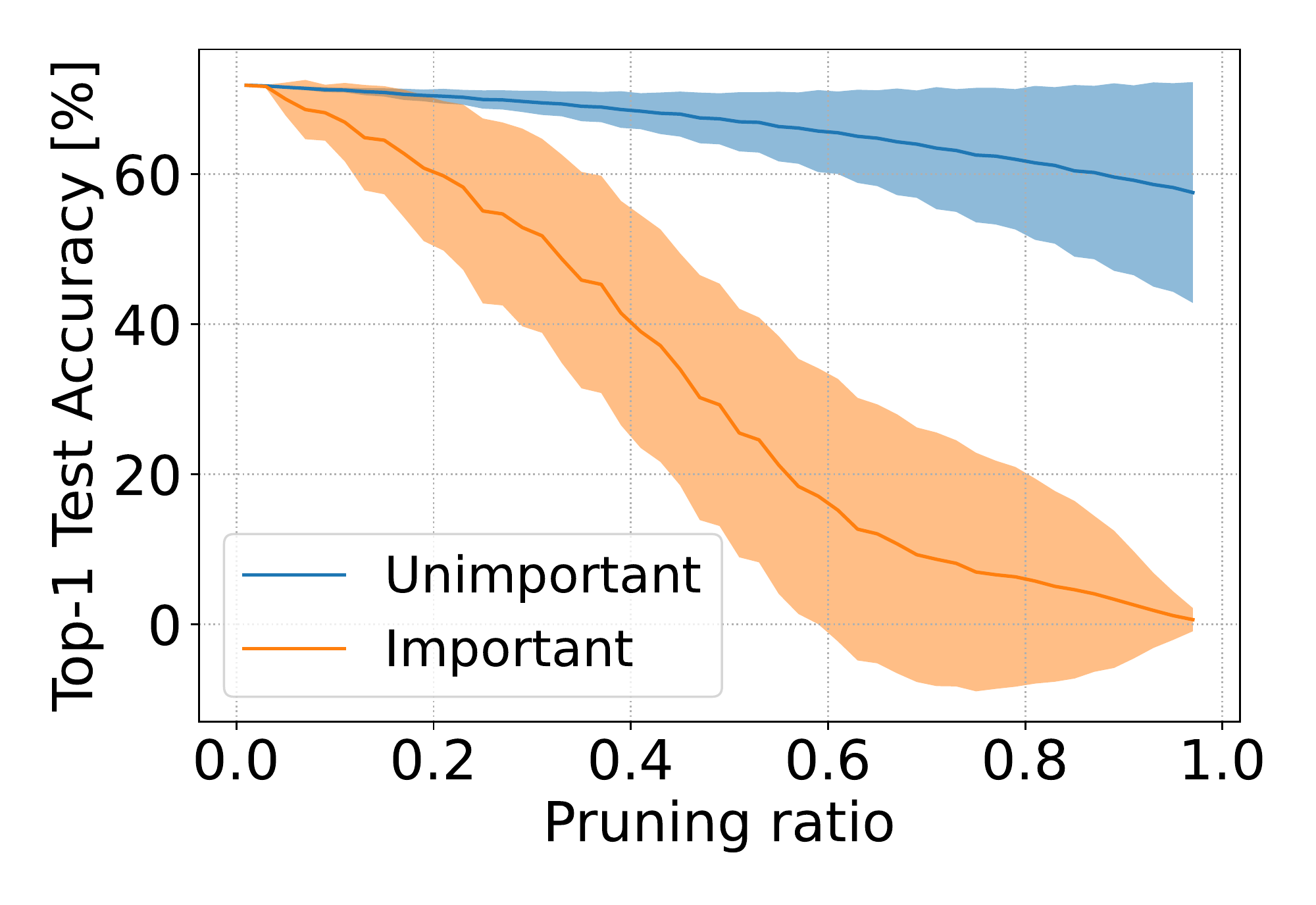}\vspace{-4mm}}
\caption{Pruning ratio of important and unimportant operations (as identified by TAS/TPS) vs. mean test accuracy on ImageNet (std. dev. shown with shade) on EfficientNet-B0, ResNet18 and MobileNet-v2. More results are given in Fig.~\ref{fig:app_acc_prune} in Appendix~\ref{app:important_pruning}. \vspace{-2mm}}
\label{fig:tips_valid_imp_op}
\end{figure*}

\subsection{Experimental Setup}
In this section, we present the following experiments: (\textit{i}) Verification of Proposition~\ref{prop:dynamic_networks}, (\textit{ii}) Validation of {TIPS}, and (\textit{iii}) Model complexity vs. accuracy results. 

For the experiments (\textit{i}) on Proposition~\ref{prop:dynamic_networks}, we build several MLP-based supernets on MNIST dataset by stacking several linear layers with 80 neurons (and adding residual connections between each two consecutive layers), then verify our Proposition~\ref{prop:dynamic_networks} on these supernets.

For the experiments (\textit{ii}) on {TIPS}, we use TPS and TAS to identify the importance of various operations in several networks (MobileNet-v2, ResNet, WideResNet, ResNext, and EfficientNet) for the ImageNet dataset. We also present the comparisons between the training convergence for our proposed {TIPS} strategy and the previous SOTA methods~\cite{chin2021joslimrudydiana,yu2019universally} with the exact same setup (\textit{i.e.}, same data augmentation, optimizer, and learning rate schedule). More training details are given in Appendix~\ref{app:hyperparam}.

Finally, for experiments (\textit{iii}), we take the MobileNet-v2 and ResNet34 trained with {TIPS} as supernets, and then search for Pareto-optimal subnetworks. We compare the accuracy-FLOPs tradeoffs of the obtained subnetworks with various training strategies.

\subsection{Verification of Proposition~\ref{prop:dynamic_networks}}
\begin{figure*}[htb]
\centering
\subfigure[Mean singular values vs. TPS]{\includegraphics[width=0.3\textwidth]{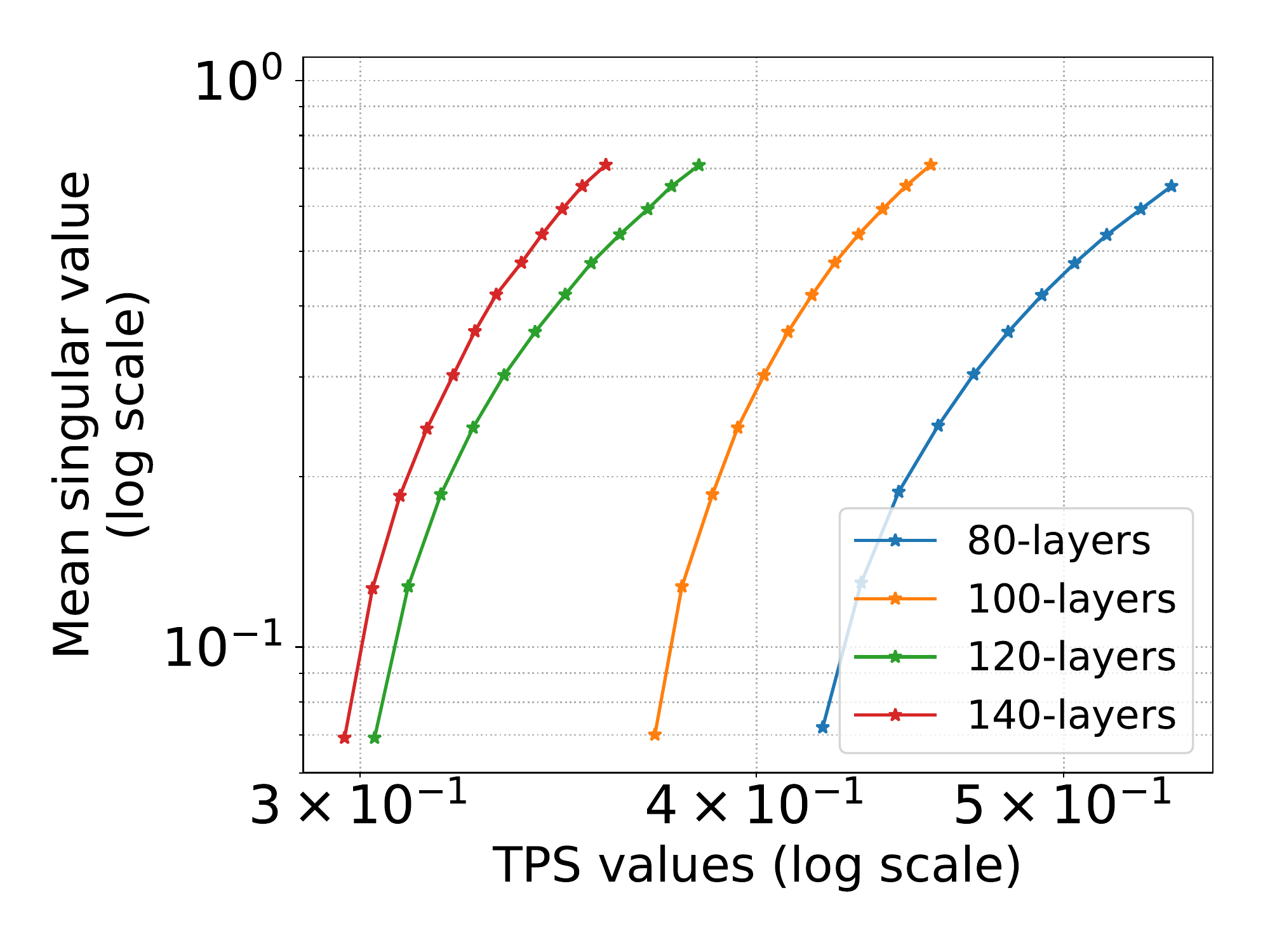}\vspace{-8mm}}
\hfill
\subfigure[MBN-v2 on ImageNet]{\includegraphics[width=0.3\textwidth]{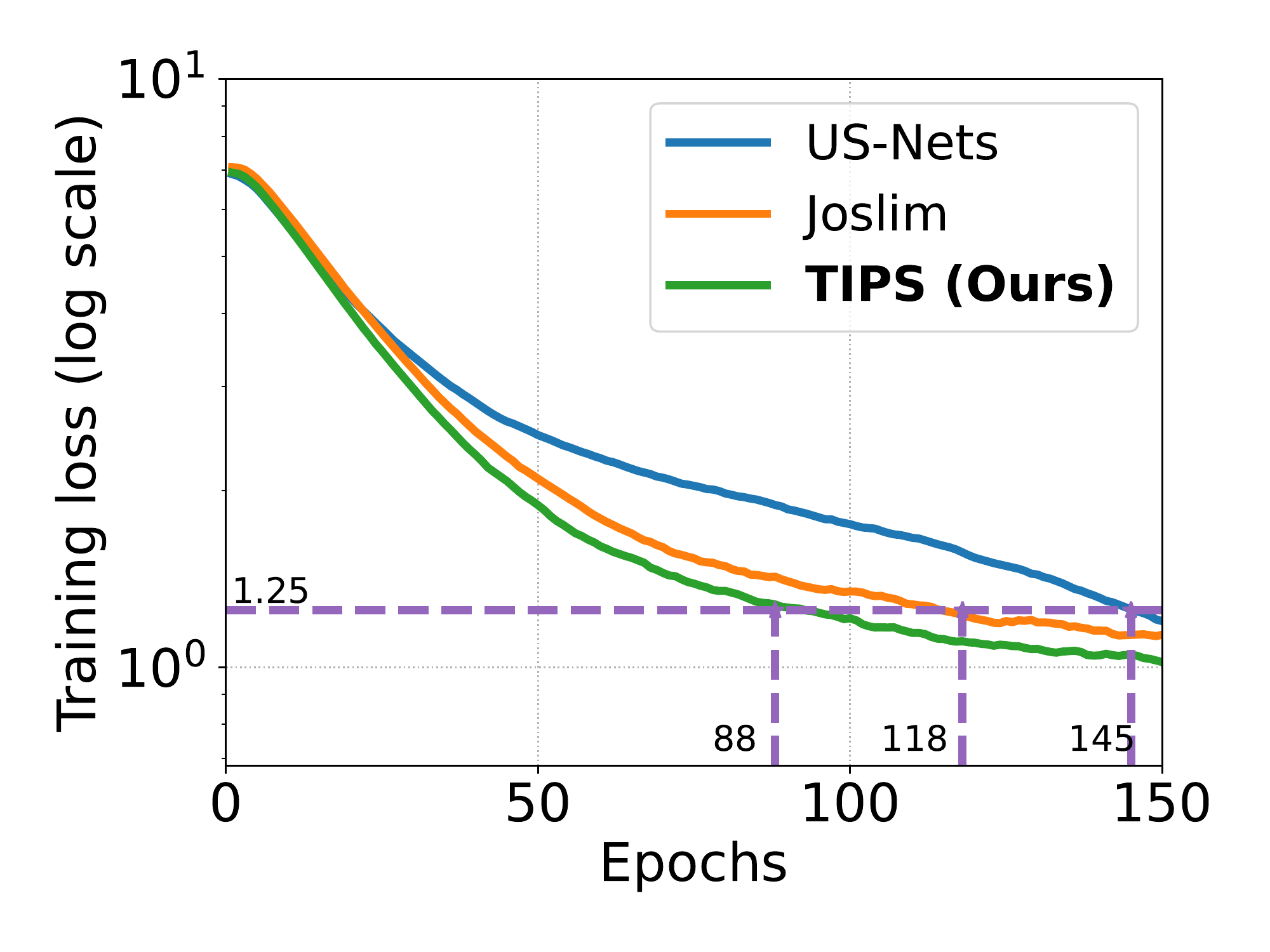}\vspace{-8mm}}
\hfill
\subfigure[MBN-v2 on Tiny-ImageNet]{\includegraphics[width=0.3\textwidth]{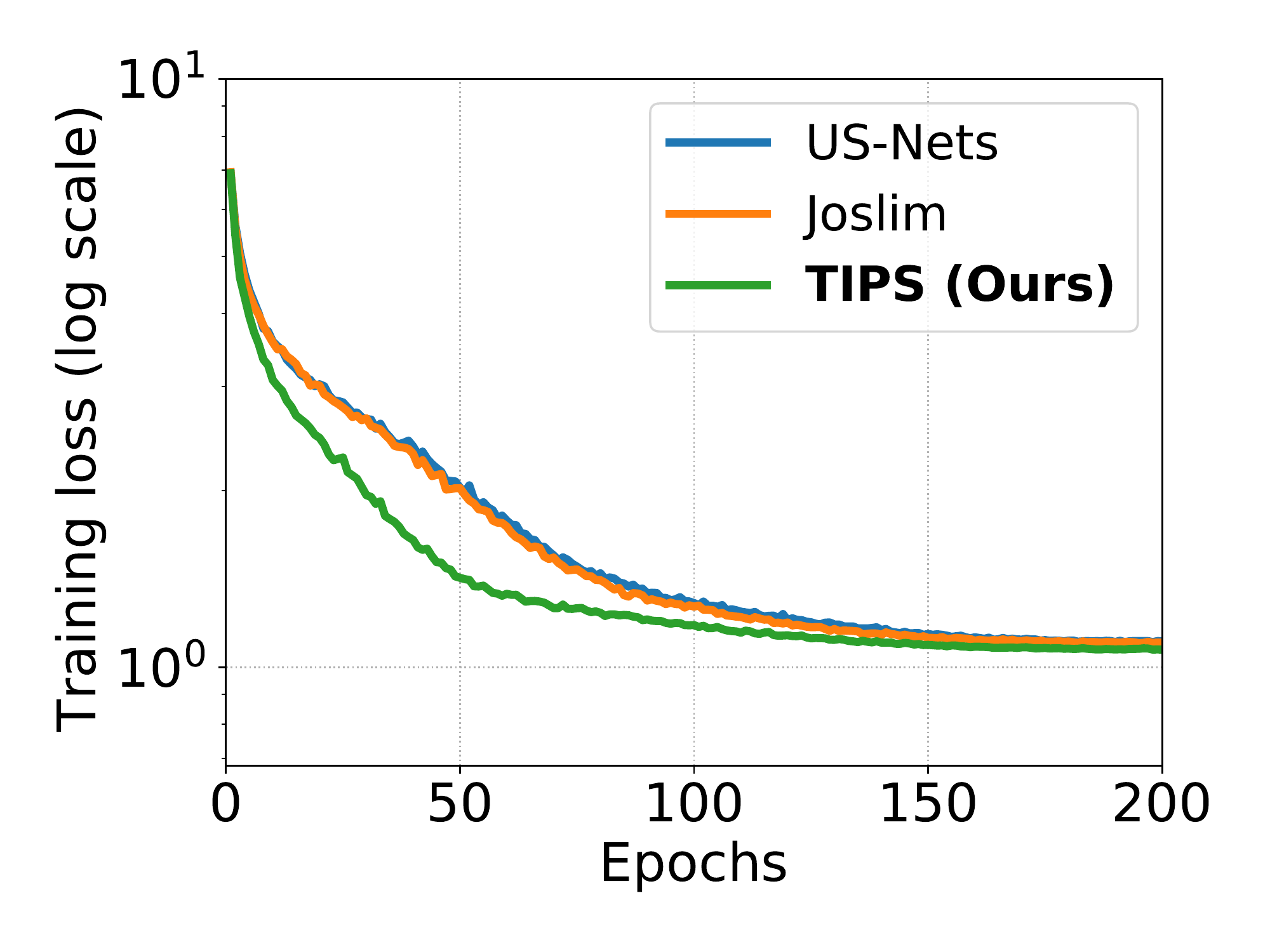}\vspace{-8mm}}
\caption{\textbf{(a)} Mean singular values ($E[\sigma]$) vs. TPS for various supernets (MLPs with various \#layers) on MNIST. Clearly, paths with higher TPS values have higher $E[\sigma]$ for a specific supernet (\textit{e.g.}, 80-layers). \textbf{(b, c)} Training loss of MobileNet-v2 (MBN-v2) based supernet vs. \#Epochs on ImageNet and Tiny-ImageNet. {TIPS} requires much fewer epochs to achieve a target training loss. For example, on ImageNet, to make the training loss less than 1.25 on ImageNet, US-Nets takes 145 epochs while {TIPS} only needs 88 epochs. \vspace{0mm}}
\label{fig:converge_bcs}
\end{figure*}

To empirically validate Proposition~\ref{prop:dynamic_networks}, we consider several supernets with 80, 100, 120, and 140 layers (along with residual connections). We then randomly sample 8 subnetworks from these supernets and use Eq.~\ref{eq:adj_mat_cases}, Eq.~\ref{eq:superadj}, and Eq.~\ref{eq:normalmatrix} to build the DTMC. After solving Eq.~\ref{eq:properyofdtmc}, we calculate the TAS for each node (\textit{i.e.}, output of various operations). Next, we set the path length to 50 as an example, then randomly sample multiple computation paths with 50 nodes from these supernets and calculate the corresponding \{TPS, mean singular value\} pairs. As shown in Fig.~\ref{fig:converge_bcs}(a), for a supernet with specific depth (\textit{e.g.}, 80 layers), higher TPS values always lead to higher mean singular values (closer to 1). These results empirically validate our Proposition~\ref{prop:dynamic_networks}.

\subsection{Validation of TIPS}\label{sec:exp_val_tips}
In order to verify the effectiveness of our topological analysis, we first explore the relationship between the \textit{important operations} and the test accuracy for various networks. To this end, before training, we first use our DTMC based framework to obtain the TAS for each node (Eq.~\ref{eq:deftas}). Next, among all the computation paths from input to output in the supernet, we find the path that has the highest TPS value (Eq.~\ref{eq:def_tps}); we mark all operations along this path as important operations. Then, we prune the output channels of each operation \textit{individually} (with pruning ratios ranging from 1\% to 99\%), without pruning the channels of any other operation in the network. Meanwhile, we measure the test accuracy of the network after each pruning step. This way, we can analyze the impact of each operation on the test accuracy of a given network. Note that, we prune the last channels first. For example, to prune a convolution layer with 64 (0-63) channels with a pruning ratio of 75\%, we directly set the output of the last 75\% (16-63) channels to zero. As shown in Fig.~\ref{fig:tips_valid_imp_op}, for various pruning ratios, pruning the important operations has a much higher accuracy drop than the unimportant ones. These experimental results show that the important operations found by our framework have a significant impact on the test accuracy of AnytimeNNs. Therefore, the proposed TAS and TPS metrics clearly identify the important operations and computation paths in AnytimeNNs. 

Next, we evaluate the impact of {TIPS} on \textit{training convergence} of the AnytimeNNs. For this experiment, we train MobileNet-v2 with width-multiplier 1.4 on ImageNet dataset with (\textit{i}) SOTA training strategies: Joslim, US-Nets, and DS-Net~\cite{chin2021joslimrudydiana,yu2019universally, li2021dynamic_dsnet}, and (\textit{ii}) our proposed {TIPS}. As explained in Section~\ref{sec:approach_nnpath}, a higher sampling probability for important operations helps more with the training process. However, if the sampling probability for important operations is too high, it hurts the diversity of sampled subnetworks. In the extreme case, we always end up sampling and training only the important operations, while the unimportant ones never get sampled and trained; this can hurt the test accuracy of AnytimeNNs. Hence, for our proposed {TIPS} strategy, we use a 50\% higher sampling probability for important operations compared to unimportant operations. For example, if 40\% of the output channels of unimportant operations are sampled, then 60\% of the output channels of important operations are sampled (since 40\%$\times$(1+0.5)=60\%). 

We note that previous methods (\textit{e.g.}, Joslim and US-Nets) use a uniform sampling for every subnetwork, \textit{i.e.}, the same sampling probability for all operations during the training process. In contrast, TIPS focuses more on important operations thus improving the LDI properties. As shown in Fig.~\ref{fig:converge_bcs}(b,c), by changing the sampling strategy, our {TIPS} based-training achieves a much faster training convergence for the supernet compared to Joslim and US-Nets. Hence, this validates that {TIPS} results in better trainability of the supernet.

\begin{table}[t]
\caption{Comparison of Top-1 test accuracy vs. FLOPS (Million [M]) with SOTA training methods on MobileNet-v2. The best results are shown with bold fonts. The results are averaged over three runs. The std. dev. values are given in Table~\ref{tab:app_mbn2_flopsacc_std} in Appendix~\ref{app:std_dev}. \vspace{0mm}}
\label{tab:mbnv2_main_flopsacc}
\begin{center}\scalebox{0.82}{
\begin{tabular}{|c|cccccc|}
\hline
\multicolumn{7}{|c|}{\textbf{CIFAR100}}\\ \hline
\multicolumn{1}{|c|}{FLOPS}                & \multicolumn{1}{c|}{20M}           & \multicolumn{1}{c|}{30M}           & \multicolumn{1}{c|}{35M}           & \multicolumn{1}{c|}{40M}           & \multicolumn{1}{c|}{45M}           & 50M           \\ \hline
\multicolumn{1}{|c|}{US-Nets}              & \multicolumn{1}{c|}{61.5}          & \multicolumn{1}{c|}{62.9}          & \multicolumn{1}{c|}{64.8}          & \multicolumn{1}{c|}{65.5}          & \multicolumn{1}{c|}{65.6}          & 66.5          \\ \hline
\multicolumn{1}{|c|}{Joslim}               & \multicolumn{1}{c|}{62.0}          & \multicolumn{1}{c|}{62.7}          & \multicolumn{1}{c|}{63.1}          & \multicolumn{1}{c|}{63.7}          & \multicolumn{1}{c|}{64.1}          & 65.0          \\ \hline
\multicolumn{1}{|c|}{DS-Net}               & \multicolumn{1}{c|}{61.8}          & \multicolumn{1}{c|}{63.8}          & \multicolumn{1}{c|}{64.8}          & \multicolumn{1}{c|}{65.3}          & \multicolumn{1}{c|}{65.5}          & 66.7          \\ \hline
\multicolumn{1}{|c|}{\textbf{TIPS (Ours)}} & \multicolumn{1}{c|}{\textbf{66.4}} & \multicolumn{1}{c|}{\textbf{66.9}} & \multicolumn{1}{c|}{\textbf{67.0}} & \multicolumn{1}{c|}{\textbf{67.6}} & \multicolumn{1}{c|}{\textbf{67.7}} & \textbf{68.2} \\ \hline\hline
\multicolumn{7}{|c|}{\textbf{Tiny-ImageNet}}\\ \hline
\multicolumn{1}{|c|}{FLOPS}                & \multicolumn{1}{c|}{80M}           & \multicolumn{1}{c|}{120M}          & \multicolumn{1}{c|}{140M}          & \multicolumn{1}{c|}{160M}          & \multicolumn{1}{c|}{180M}          & 200M          \\ \hline
\multicolumn{1}{|c|}{US-Nets}              & \multicolumn{1}{c|}{47.0}          & \multicolumn{1}{c|}{47.3}          & \multicolumn{1}{c|}{48.3}          & \multicolumn{1}{c|}{49.0}          & \multicolumn{1}{c|}{50.2}          & 51.4          \\ \hline
\multicolumn{1}{|c|}{Joslim}               & \multicolumn{1}{c|}{47.4}          & \multicolumn{1}{c|}{47.9}          & \multicolumn{1}{c|}{48.7}          & \multicolumn{1}{c|}{49.5}          & \multicolumn{1}{c|}{50.3}          & 50.7          \\ \hline
\multicolumn{1}{|c|}{DS-Net}               & \multicolumn{1}{c|}{46.9}          & \multicolumn{1}{c|}{47.4}          & \multicolumn{1}{c|}{48.1}          & \multicolumn{1}{c|}{48.7}          & \multicolumn{1}{c|}{50.3}          & 50.8          \\ \hline
\multicolumn{1}{|c|}{\textbf{TIPS (Ours)}} & \multicolumn{1}{c|}{\textbf{53.5}} & \multicolumn{1}{c|}{\textbf{53.8}} & \multicolumn{1}{c|}{\textbf{54.0}} & \multicolumn{1}{c|}{\textbf{54.4}} & \multicolumn{1}{c|}{\textbf{54.9}} & \textbf{55.1} \\ \hline\hline
\multicolumn{7}{|c|}{\textbf{ImageNet}}\\ \hline
\multicolumn{1}{|c|}{FLOPS}                & \multicolumn{1}{c|}{260M}          & \multicolumn{1}{c|}{320M}          & \multicolumn{1}{c|}{400M}          & \multicolumn{1}{c|}{450M}          & \multicolumn{1}{c|}{500M}          & 600M          \\ \hline
\multicolumn{1}{|c|}{US-Nets}              & \multicolumn{1}{c|}{70.6}          & \multicolumn{1}{c|}{71.6}          & \multicolumn{1}{c|}{71.8}          & \multicolumn{1}{c|}{72.1}          & \multicolumn{1}{c|}{72.3}          & 72.9          \\ \hline
\multicolumn{1}{|c|}{Joslim}               & \multicolumn{1}{c|}{70.8}          & \multicolumn{1}{c|}{71.9}          & \multicolumn{1}{c|}{72.5}          & \multicolumn{1}{c|}{72.7}          & \multicolumn{1}{c|}{72.9}          & 73.4          \\ \hline
\multicolumn{1}{|c|}{DS-Net}               & \multicolumn{1}{c|}{70.6}          & \multicolumn{1}{c|}{72.1}          & \multicolumn{1}{c|}{72.5}          & \multicolumn{1}{c|}{72.6}          & \multicolumn{1}{c|}{73.0}          & 73.3          \\ \hline
\multicolumn{1}{|c|}{\textbf{TIPS (Ours)}} & \multicolumn{1}{c|}{\textbf{71.8}} & \multicolumn{1}{c|}{\textbf{73.2}} & \multicolumn{1}{c|}{\textbf{73.6}} & \multicolumn{1}{c|}{\textbf{74.0}} & \multicolumn{1}{c|}{\textbf{74.3}} & \textbf{74.7} \\ \hline
\end{tabular}
}
\vspace{0mm}
\end{center}
\end{table}

\begin{table}[t]
\caption{Comparison of Top-1 test accuracy vs. FLOPS (Million/Giga [M/G]) with SOTA training methods on ResNet34. The best results are shown with bold fonts. The results are averaged over three runs. The std. dev. values are shown in Table~\ref{tab:app_rn34_flopsacc_std} in Appendix~\ref{app:std_dev}. \vspace{0mm}}
\label{tab:rn34_main_flopsacc}
\begin{center}\scalebox{0.82}{
\begin{tabular}{|ccccccc|}
\hline
\multicolumn{7}{|c|}{\textbf{CIFAR100}}\\ \hline
\multicolumn{1}{|c|}{FLOPS}                & \multicolumn{1}{c|}{120M}          & \multicolumn{1}{c|}{180M}          & \multicolumn{1}{c|}{200M}          & \multicolumn{1}{c|}{220M}          & \multicolumn{1}{c|}{240M}          & 260M          \\ \hline
\multicolumn{1}{|c|}{US-Nets}              & \multicolumn{1}{c|}{63.1}          & \multicolumn{1}{c|}{63.9}          & \multicolumn{1}{c|}{64.4}          & \multicolumn{1}{c|}{64.8}          & \multicolumn{1}{c|}{65.0}          & 65.4          \\ \hline
\multicolumn{1}{|c|}{Joslim}               & \multicolumn{1}{c|}{65.8}          & \multicolumn{1}{c|}{66.2}          & \multicolumn{1}{c|}{66.7}          & \multicolumn{1}{c|}{67.0}          & \multicolumn{1}{c|}{67.3}          & 67.4          \\ \hline
\multicolumn{1}{|c|}{DS-Net}               & \multicolumn{1}{c|}{64.4}          & \multicolumn{1}{c|}{65.9}          & \multicolumn{1}{c|}{66.2}          & \multicolumn{1}{c|}{66.4}          & \multicolumn{1}{c|}{66.5}          & 66.6          \\ \hline
\multicolumn{1}{|c|}{\textbf{TIPS (Ours)}} & \multicolumn{1}{c|}{\textbf{67.3}} & \multicolumn{1}{c|}{\textbf{67.4}} & \multicolumn{1}{c|}{\textbf{67.8}} & \multicolumn{1}{c|}{\textbf{67.9}} & \multicolumn{1}{c|}{\textbf{68.1}} & \textbf{68.2} \\ \hline\hline
\multicolumn{7}{|c|}{\textbf{Tiny-ImageNet}}\\ \hline
\multicolumn{1}{|c|}{FLOPS}                & \multicolumn{1}{c|}{130M}          & \multicolumn{1}{c|}{190M}          & \multicolumn{1}{c|}{220M}          & \multicolumn{1}{c|}{250M}          & \multicolumn{1}{c|}{270M}          & 300M          \\ \hline
\multicolumn{1}{|c|}{US-Nets}              & \multicolumn{1}{c|}{42.9}          & \multicolumn{1}{c|}{43.2}          & \multicolumn{1}{c|}{44.3}          & \multicolumn{1}{c|}{44.7}          & \multicolumn{1}{c|}{44.9}          & 45.2          \\ \hline
\multicolumn{1}{|c|}{Joslim}               & \multicolumn{1}{c|}{44.9}          & \multicolumn{1}{c|}{\textbf{45.0}} & \multicolumn{1}{c|}{45.3}          & \multicolumn{1}{c|}{45.4}          & \multicolumn{1}{c|}{45.5}          & 45.8          \\ \hline
\multicolumn{1}{|c|}{DS-Net}               & \multicolumn{1}{c|}{41.8}          & \multicolumn{1}{c|}{43.0}          & \multicolumn{1}{c|}{43.8}          & \multicolumn{1}{c|}{43.9}          & \multicolumn{1}{c|}{44.1}          & 44.2          \\ \hline
\multicolumn{1}{|c|}{\textbf{TIPS (Ours)}} & \multicolumn{1}{c|}{\textbf{44.1}} & \multicolumn{1}{c|}{44.6}          & \multicolumn{1}{c|}{\textbf{45.4}} & \multicolumn{1}{c|}{\textbf{45.8}} & \multicolumn{1}{c|}{\textbf{45.9}} & \textbf{46.0} \\ \hline\hline
\multicolumn{7}{|c|}{\textbf{ImageNet}}\\ \hline
\multicolumn{1}{|c|}{FLOPS}                & \multicolumn{1}{c|}{1.5G}          & \multicolumn{1}{c|}{2.2G}          & \multicolumn{1}{c|}{2.8G}          & \multicolumn{1}{c|}{3.0G}          & \multicolumn{1}{c|}{3.2G}          & 3.6G          \\ \hline
\multicolumn{1}{|c|}{US-Nets}              & \multicolumn{1}{c|}{67.8}          & \multicolumn{1}{c|}{69.2}          & \multicolumn{1}{c|}{69.7}          & \multicolumn{1}{c|}{70.1}          & \multicolumn{1}{c|}{70.2}          & 70.5          \\ \hline
\multicolumn{1}{|c|}{Joslim}               & \multicolumn{1}{c|}{68.0}          & \multicolumn{1}{c|}{\textbf{69.4}} & \multicolumn{1}{c|}{69.6}          & \multicolumn{1}{c|}{70.0}          & \multicolumn{1}{c|}{70.2}          & 70.4          \\ \hline
\multicolumn{1}{|c|}{DS-Net}               & \multicolumn{1}{c|}{66.0}          & \multicolumn{1}{c|}{67.0}          & \multicolumn{1}{c|}{68.8}          & \multicolumn{1}{c|}{69.4}          & \multicolumn{1}{c|}{69.9}          & 70.0          \\ \hline
\multicolumn{1}{|c|}{\textbf{TIPS (Ours)}} & \multicolumn{1}{c|}{\textbf{68.4}} & \multicolumn{1}{c|}{69.3}          & \multicolumn{1}{c|}{\textbf{70.8}} & \multicolumn{1}{c|}{\textbf{71.1}} & \multicolumn{1}{c|}{\textbf{71.4}} & \textbf{71.9} \\ \hline
\end{tabular}
}\vspace{0mm}
\end{center}
\end{table}

\subsection{Pareto-Optimal AnytimeNN Search}\label{sec:pods}
We use the Algorithm~\ref{alg:paretosearch} to search for Pareto-optimal subnetworks under various hardware constraints for MobileNet-v2 and ResNet34. After the search, we evaluate the obtained Pareto-optimal subnetworks and get their real test accuracy.

Table~\ref{tab:mbnv2_main_flopsacc} demonstrates that our proposed {TIPS} achieves significantly higher accuracy than SOTA given similar FLOPs for ImageNet on MobileNet-v2. For example, assuming the hardware constraint is 500M FLOPs, {TIPS} has a 1.4\%-2\% higher accuracy on ImageNet than the SOTA; on Tiny-ImageNet with an 80M FLOPs budget, {TIPS} has 6.1\%-6.6\% higher accuracy than SOTA.

Moreover, as shown in Table~\ref{tab:rn34_main_flopsacc}, given the ResNet34 supernet with a 3.6G FLOPs budget on ImageNet, {TIPS} achieves 1.4\%, 1.5\%, and 1.9\% higher test accuracy than Joslim, US-Nets and DS-Net, respectively~\cite{chin2021joslimrudydiana,yu2019universally,li2021dynamic_dsnet}. On CIFAR100 dataset with a 120M FLOPs budget, {TIPS} has 1.5\%, 2.9\%, and 4.2\% higher accuracy Joslim, US-Nets and DS-Net, respectively.

\begin{table}[t]
\caption{Top-1 test accuracy vs. latency of MobileNet-v2 on ImageNet for RaspberryPi-3B+. The results are averaged over three runs. \vspace{0mm}}
\label{tab:latency_acc}
\begin{center}\scalebox{0.8}{
\begin{tabular}{|c|c||ccccc|}
\hline
Method                        & Metric        & \multicolumn{5}{c|}{Results}                                                                                                \\ \hline\hline
\multirow{2}{*}{Joslim}       & Latency (ms): & \multicolumn{1}{c|}{176}  & \multicolumn{1}{c|}{232}  & \multicolumn{1}{c|}{305}  & \multicolumn{1}{c|}{341}  & 406  \\ \cline{2-7} 
                              & Accuracy (\%) & \multicolumn{1}{c|}{70.8} & \multicolumn{1}{c|}{71.9} & \multicolumn{1}{c|}{72.5} & \multicolumn{1}{c|}{72.9} & 73.4 \\ \hline\hline
\multirow{2}{*}{\textbf{TIPS  (Ours)}} & Latency (ms): & \multicolumn{1}{c|}{190}  & \multicolumn{1}{c|}{245}  & \multicolumn{1}{c|}{298}  & \multicolumn{1}{c|}{362}  & 413  \\ \cline{2-7} 
                              & Accuracy (\%) & \multicolumn{1}{c|}{71.8} & \multicolumn{1}{c|}{73.2} & \multicolumn{1}{c|}{73.6} & \multicolumn{1}{c|}{74.3} & 74.7 \\ \hline
\end{tabular}
}\vspace{0mm}
\end{center}
\end{table}

\textbf{Latency vs. Accuracy Trade-off} Besides FLOPs vs. accuracy, we also compare the latency vs. accuracy tradeoff of subnetworks obtained by TIPS and Joslim. As shown in Table~\ref{tab:latency_acc}, {TIPS} achieves higher accuracy than Joslim, given a similar latency. For example, assuming the latency constraint is around 300ms, {TIPS} has a 1.1\% higher accuracy on ImageNet than the Joslim.

\begin{figure}[htb]
\vspace{0mm}
\centering
\includegraphics[width=0.45\textwidth]{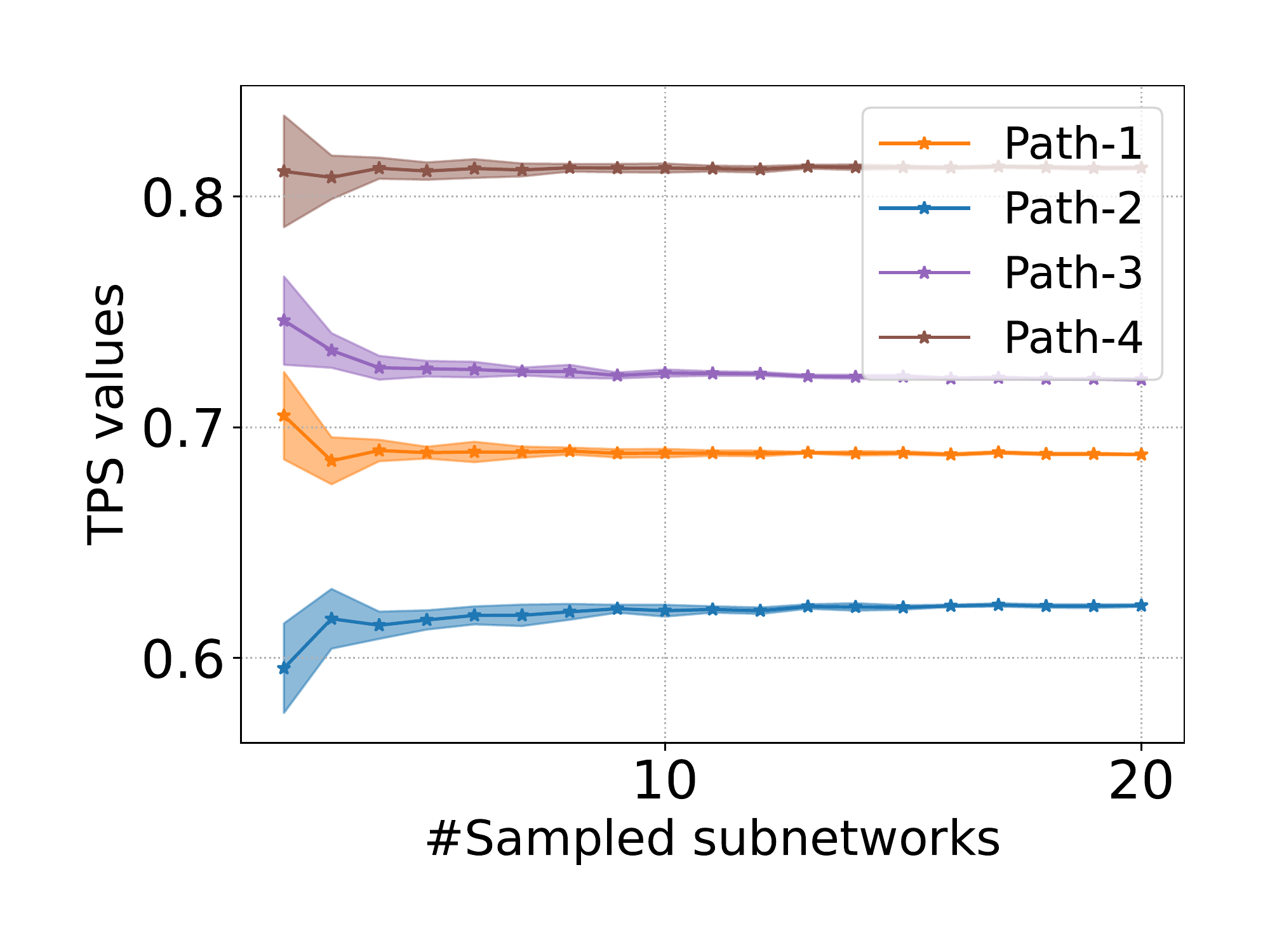}\vspace{-6mm}
\caption{Stability of TPS values vs. \#subnetworks over 10 runs (std. dev. shown with shade) for MobileNet-v2. The variation is negligible when \#subnetworks is larger than 4.}\vspace{-2mm}
\label{fig:ablation_num_net}
\end{figure}

\begin{figure}[htb]
\vspace{0mm}
\centering
\includegraphics[width=0.45\textwidth]{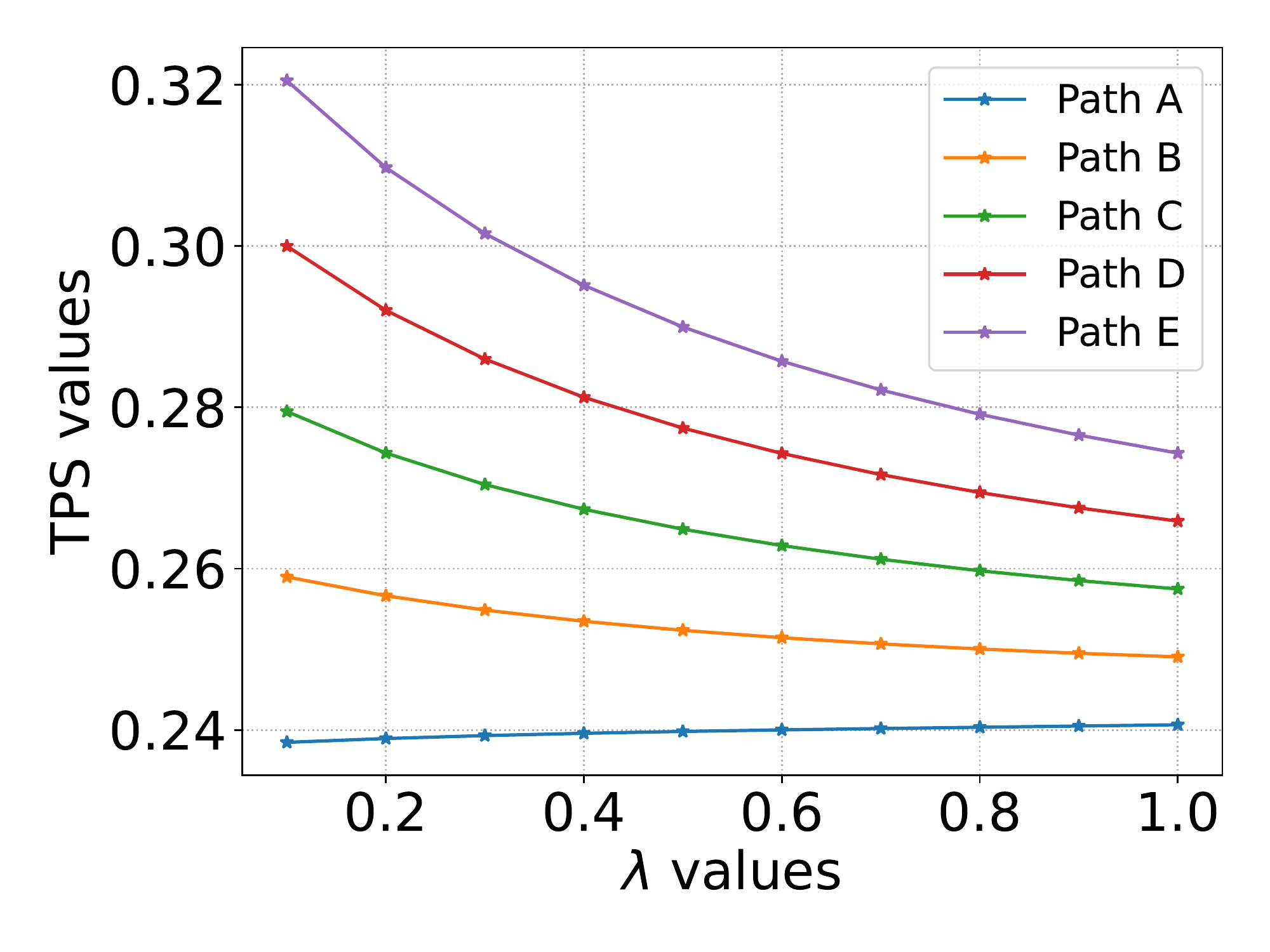}\vspace{-6mm}
\caption{TPS values vs. $\lambda$ values for MobileNet-v2. The ranking among different paths remains the same for various $\lambda$ values.}
\label{fig:ablation_lambda}
\end{figure}

\subsection{Ablation Studies}\label{sec:exp_ablation}

\noindent\textbf{Stability of TPS analysis}
We vary the \#sampled subnetworks from 2 to 20 for MobileNet-v2. As shown in Fig.~\ref{fig:ablation_num_net}, four subnetworks are typically enough to make the TPS value converge. In practice, we sample 8 subnetworks and the standard deviation of TPS values is less than 2.5\% of the mean values. 

\noindent\textbf{Effect of $\bm{\mathbf{\lambda}}$} Finally, we fix the \#sampled subnetworks to 8 and vary the $\lambda$ value in the hyper-adjacency matrix (Eq.~\ref{eq:superadj}) from 0.1 to 1 for MobileNet-v2. As shown in Fig.~\ref{fig:ablation_lambda}, the ranking among different paths remains the same under various $\lambda$ values. Therefore, our approach is robust to $\lambda$ values variation. In our approach, we set the value of $\lambda$ to `1'. We discuss this in Appendix~\ref{app:detail_tips}. 

\subsection{Limitations and Future Work}
Our current framework (TIPS) has been primarily verified on CNNs with variable width and depth. We plan to explore it with other AnytimeNNs (\textit{e.g.}, multi-branch and early-exit networks) and other types of networks (\textit{e.g.}, transformers and graph neural networks). Also, in the current version, the hardware constraints are considered \textit{after} the supernet training; we intend to consider incorporating hardware awareness into the training process as well.
\section{Conclusion}\label{sec:conclusion}
In this work, we have proposed a new methodology to \textit{automatically design the AnytimeNNs} under various hardware budgets. To this end, by modeling the training process of AnytimeNNs as a DTMC, we have proposed two metrics -- TAS and TPS -- to characterize the important operations in AnytimeNNs. We have shown that these important operations and computation paths significantly impact the accuracy of AnytimeNNs. Based on this, we have proposed a new training method called \textit{TIPS}. Experimental results show that {TIPS} has a faster training convergence speed than SOTA training methods for anytime inference. Our experimental results demonstrate that our framework can achieve SOTA accuracy-FLOPs trade-offs, while achieving 2\%-6.6\% accuracy improvements on CIFAR100, Tiny-ImageNet and ImageNet datasets compared to existing approaches for anytime inference. 

\section*{Acknowledgement}
This work was supported in part by the US National Science Foundation (NSF) grants CNS-2007284 and CCF-2107085.

\bibliography{example_paper}
\bibliographystyle{icml2023}

\newpage
\appendix
\onecolumn
\newpage
\appendix
\onecolumn
\section{Supplementary Results for Importance Analysis}\label{app:important_pruning}
\begin{figure}[h]
\vspace{0mm}
  \subfigure[WideResNet50-2]{
	   \centering
	   \includegraphics[width=0.32\textwidth]{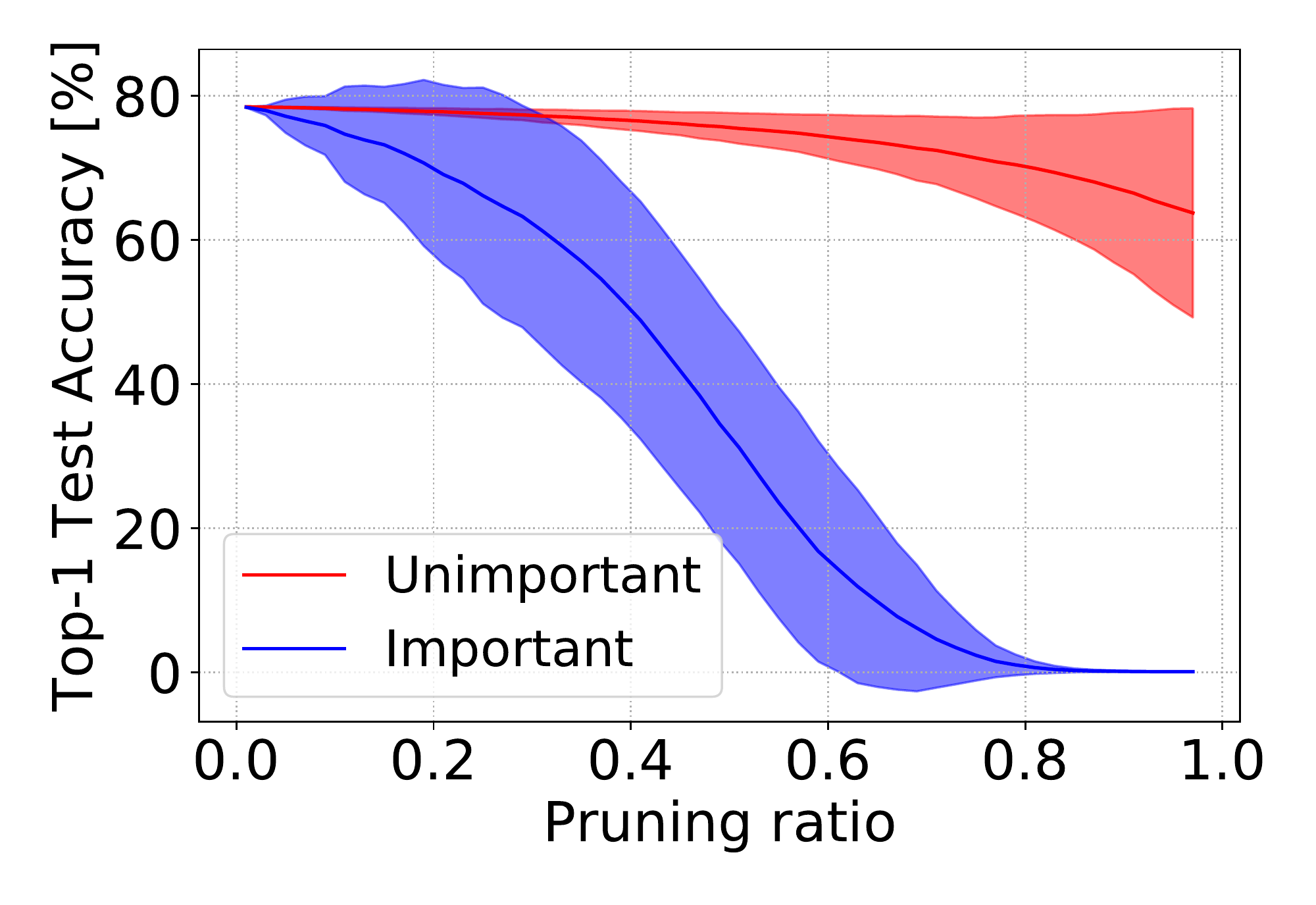}}
 \hfill
  \subfigure[ResNet50]{

	   \centering
	   \includegraphics[width=0.32\textwidth]{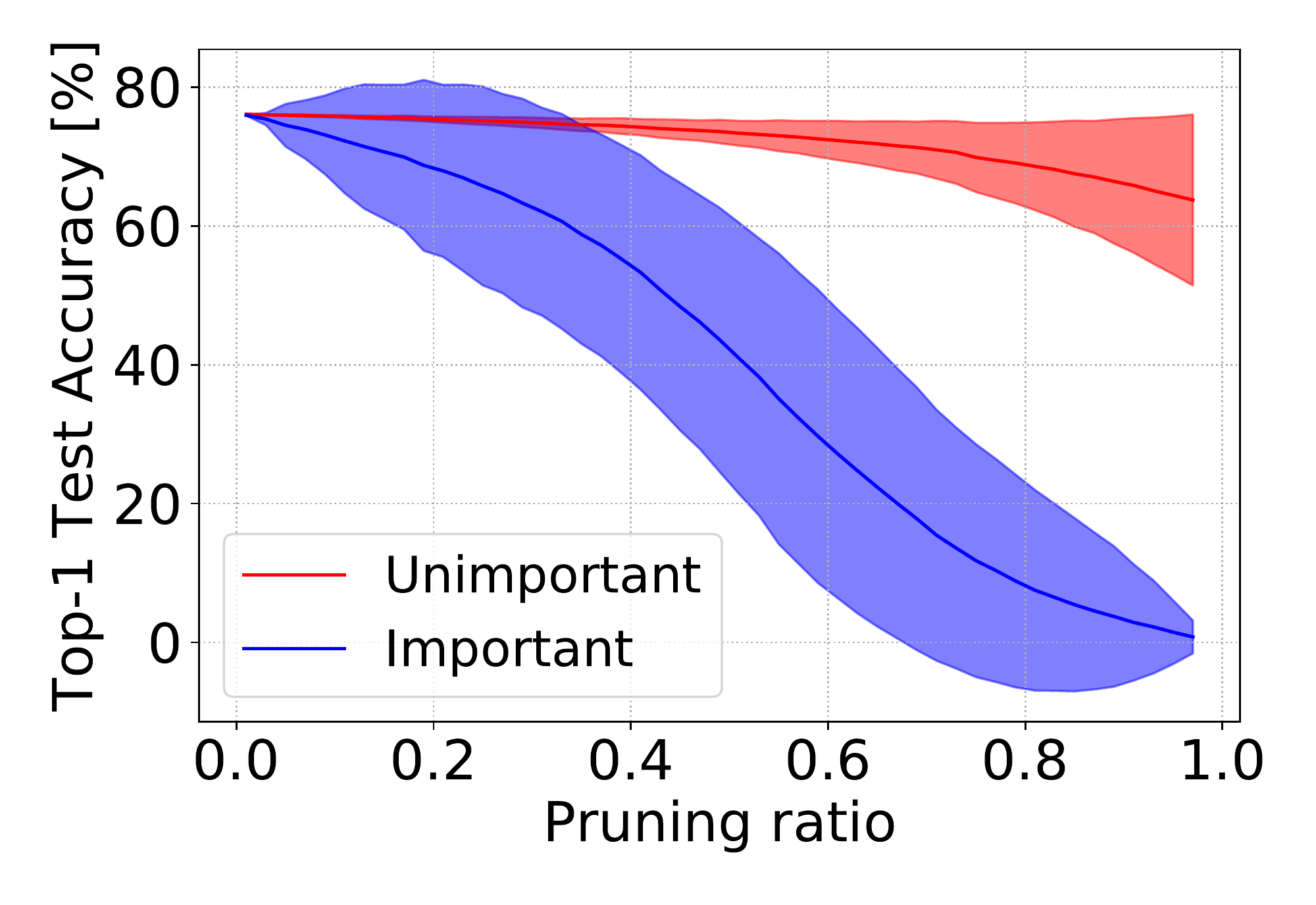}}
 \hfill	
  \subfigure[ResNext50-32x4d]{
	   \centering
	   \includegraphics[width=0.32\textwidth]{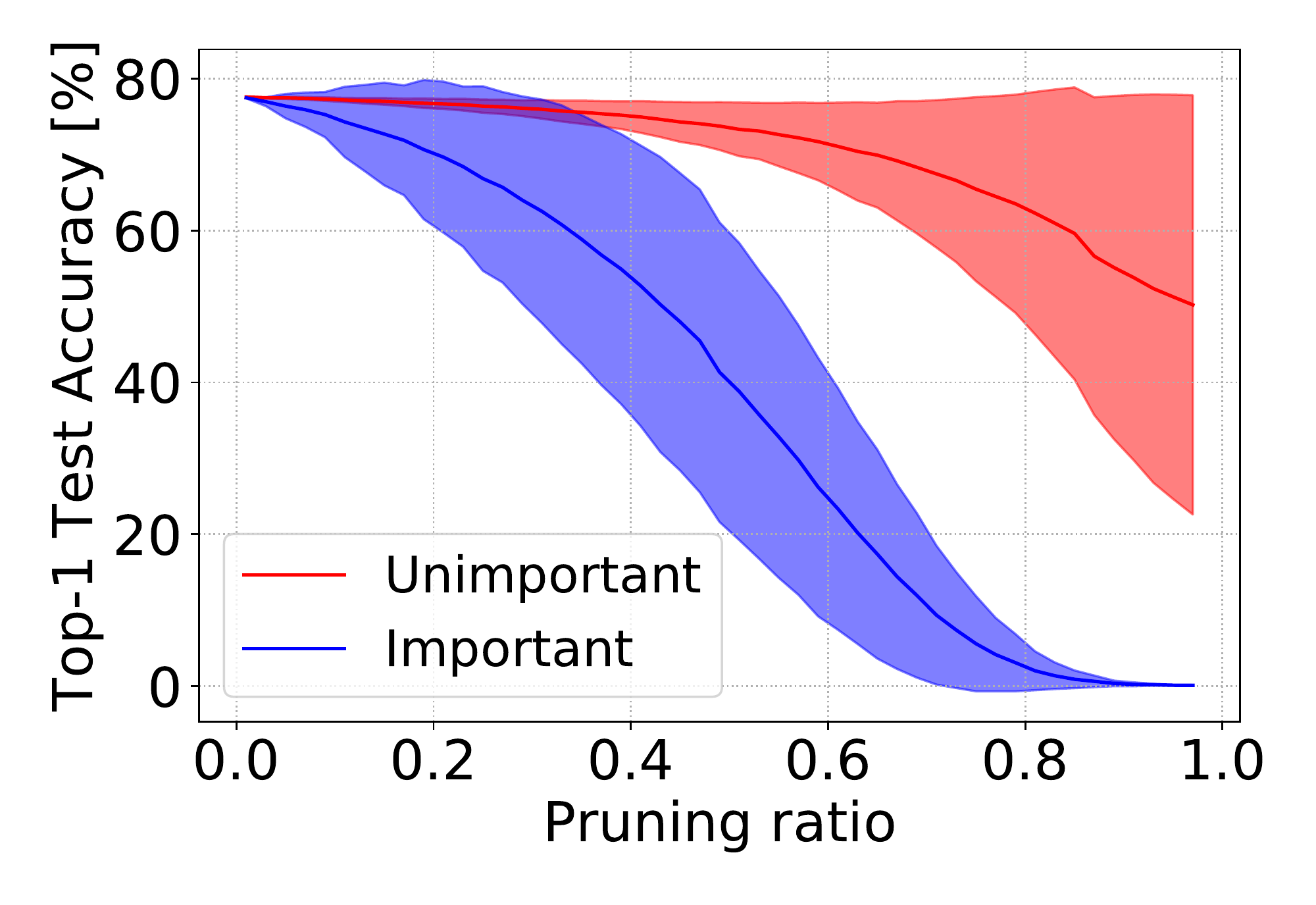}}
\caption{Pruning ratio of important and unimportant operations vs. mean test accuracy on ImageNet (standard deviations are drawn with shade). We prune the output of each operation with various pruning ratios and obtain the test accuracy. We calculate the mean accuracy under various pruning ratios for important operations and unimportant operations. As shown, for all networks, given the same pruning ratio, important operations have a much higher impact on accuracy than unimportant ones.}
\label{fig:app_acc_prune}
\end{figure}

The plots below (Fig.~\ref{fig:app_acc_prune}) supplement the results in Fig.~\ref{fig:tips_valid_imp_op} in the main paper.
As shown in Fig.~\ref{fig:app_acc_prune}, for various pruning ratios, pruning the important operations has a much higher accuracy drop than the unimportant ones. These experimental results show that the important operations found by our framework have a significant impact on the test accuracy of AnytimeNNs. 

\section{Details of the Training Methods}
\subsection{Construction of DTMC for AnytimeNNs}\label{app:dtmc}
An irreducible, aperiodic, and homogeneous DTMC has a unique state stationary distribution ($\bm{\pi}$)~\cite{hajek2015randombookmarkov}.
We analyze the above three requirements (irreducibility, aperiodicity, and homogeneity) for our problem as follows:

\textbf{Irreducibility} To ensure the constructed DTMC is irreducible, and following a similar idea from PageRank~\cite{berkhin2005surveypagerank}, we add a small transition probability $\kappa$ between each pair of states to the original $\hat{\bm{H}}(\lambda)$ (Eq.~\ref{eq:superadj} in the main paper) as follows:
\begin{equation}\label{eq:modify_h}
    \tilde{\bm{H}}(\lambda) =(1-\kappa)\hat{\bm{H}}(\lambda) +\kappa\bm{U}
\end{equation}
where $\bm{U}$ is a all-one matrix with all elements equal to `1'. Next, we use the slightly modified $\tilde{\bm{H}}(\lambda)$ to construct the DTMC. Hence, the introduced transition probability $\kappa$ guarantees that every two states in the DTMC are accessible to each other with a probability at least $\kappa$. As such, we ensure that the DTMC constructed by $\tilde{\bm{H}}(\lambda)$ is always irreducible.  
In practice, we set the value of $\kappa$ very small (\textit{e.g.}, $\kappa=10^{-5}$) to minimize the impact of the introduced transition probability.

\textbf{Aperiodicity} According to~\cite{hajek2015randombookmarkov}, for a given DTMC, if it is irreducible and there exist some self-loop transition among its states, then the DTMC is a aperiodic DTMC. (\textit{i}) The modified $\tilde{\bm{H}}(\lambda)$ in the above discussion already ensures the DTMC is irreducible. (\textit{ii}) Recall that in Eq.~\ref{eq:adj_mat_cases} (in the main paper), when we build the adjacency matrix $\bm{A}_k$, we set values of $\bm{A}_k(1,1)$ and $\bm{A}_k(N,N)$ as `1'. Hence, there are self-loops for the first and last states (\textit{i.e.}, nodes) of each sub-networks. These two conditions ensure the constructed DTMC is also aperiodic. 

\textbf{Homogeneity}
For a given DTMC, if the probabilities of state transitions are independent of time, then the DTMC is a homogeneous DTMC. In our case, the probability of state transition is determined by the sampling strategy. Intuitively, if the sampling strategy remains the same over time, then the probabilities of state transitions are the same for different time moments. Hence, in this work, it is reasonable to assume that the constructed DTMC is homogeneous as well.

In summary, by ensuring the irreducibility, aperiodicity and the assumption of homogeneity of our constructed DTMC, we can always find the stationary state distribution $\bm{\pi}$ and use it to conduct the TAS and TPS analysis.

\subsection{Training Hyperparameters}\label{app:hyperparam}
We use the SGD with a momentum of 0.9 as the optimizer and set the initial learning rate as 0.04. We set the batch-size as 512 and train the MobileNet-v2 for 150 epochs with a cosine annealing learning rate schedule on ImageNet dataset. We train the ResNet-34 for 90 epochs with the same optimizer, batch-size, and learning rate schedule on ImageNet dataset. When we train the MobileNet-v2 and ResNet-34 on CIFAR100 and Tiny-ImageNet datasets, we use the same optimizer; we reduce the batch-size to 256 and train these networks for 200 epochs with the initial learning rate as 0.08 and a cosine annealing learning rate schedule. 

\noindent\textbf{Loss function}
For each training step, we randomly sample three sub-networks $G_k,\ k=1,2,3$. In practice, to further increase the diversity of sub-networks, we conduct the sampling process at a finer level of granularity, \textit{i.e.}, at \textit{channel-level}. For example, in MobileNet-v2, we found that the layers within the block with stride=2 are important operations. Consequently, for each sub-network, we sample each channel with a 50\% higher probability for these important operations compared to the channels that correspond to the unimportant operations. 

Overall, we use the cross entropy loss together with the knowledge distillation function to train the AnytimeNNs for all these baseline methods and {TIPS}. For the same batch of input images, we combine these three subnetworks $G_k,\ k=1,2,3$ as well as the entire supernet $G$, as follows:
\begin{equation}
    Loss =\sum_{Net \in \{G, G_1,G_2,G_3\}}\mathcal{L_{CE}}(y, Net(x)) + \sum_{Net \in \{G_1,G_2,G_3\}}\mathcal{L_{KD}}(Net(x), G(k))
\end{equation}
where $x$, $y$, $\mathcal{L_{CE}}$ and $\mathcal{L_{KD}}$ are the input batch of images, labels, cross-entropy loss function and distillation function, respectively.
In our work, the distillation function $\mathcal{L_{KD}}$ is the same as the one used in~\cite{chin2021joslimrudydiana}.

\begin{table}[h]
\caption{Comparison of Top-1 test accuracy vs. FLOPS (in millions [M]) with SOTA training methods on MobileNet-v2. Best results are shown with bold fonts. Results are averaged over three runs. \vspace{0mm}}
\label{tab:app_mbn2_flopsacc_std}
\begin{center}\scalebox{0.95}{

\begin{tabular}{|c|c||c|c|c|c|c|c|}
\hline
\multirow{5}{*}{CIFAR100}      & FLOPS         & 20M                   & 30M                                  & 35M                                  & 40M                                & 45M                                  & 50M                                  \\ \cline{2-8} 
                               & US-Nets~\cite{yu2019universally} & 61.5$\pm$0.4          & 62.9$\pm$0.6                         & 64.8$\pm$0.3                         & 65.5$\pm$0.3                       & 65.6$\pm$0.1                         & 66.5$\pm$0.1                         \\ \cline{2-8} 
                               & Joslim~\cite{chin2021joslimrudydiana}  & 62.0$\pm$0.4          & 62.7$\pm$0.4                         & 63.1$\pm$0.3                        & 63.7$\pm$0.2                       & 64.1$\pm$0.3                         & 65.0$\pm$0.2                           \\ \cline{2-8} 
                               & DS-Net~\cite{li2021dynamic_dsnet}         & 61.8$\pm$0.6          & 63.8$\pm$0.3                         & 64.8$\pm$0.2                         & 65.3$\pm$0.2                       & 65.5$\pm$0.3                         & 66.7$\pm$0.2                         \\ \cline{2-8} 
                               & \textbf{TIPS}   & \textbf{66.4$\pm$0.5} & \textbf{66.9$\pm$0.1}                & \textbf{67.0$\pm$0.1}                & \textbf{67.6$\pm$0.3}              & \textbf{67.7$\pm$0.1}                & \textbf{68.2$\pm$0.3}                \\ \hline\hline
\multirow{5}{*}{Tiny-ImageNet} & FLOPS         & 80M                   & 120M                                 & 140M                                 & 160M                               & 180M                                 & 200M                                 \\ \cline{2-8} 
                               & US-Nets~\cite{yu2019universally} & 47.0$\pm$0.5          & 47.3$\pm$0.1                         & 48.3$\pm$0.3                         & 49.0$\pm$0.1                       & 50.2$\pm$0.3                         & 51.4$\pm$0.2                         \\ \cline{2-8} 
                               & Joslim~\cite{chin2021joslimrudydiana}  & 47.4$\pm$0.4          & 47.9$\pm$0.4                         & 48.7$\pm$0.1                         & 49.5$\pm$0.2                       & 50.3$\pm$0.3                         & 50.7$\pm$0.4                         \\ \cline{2-8} 
                               & DS-Net~\cite{li2021dynamic_dsnet}         & 46.9$\pm$0.3          & 47.4$\pm$0.3                         & 48.1$\pm$0.2                         & 48.7$\pm$0.1                       & 50.3$\pm$0.2                         & 50.8$\pm$0.2                         \\ \cline{2-8} 
                               & \textbf{TIPS}   & \textbf{53.5$\pm$0.3} & \textbf{53.8$\pm$0.2}                & \textbf{54.0$\pm$0.1}                & \textbf{54.4$\pm$0.3}              & \textbf{54.9$\pm$0.2}                & \textbf{55.1$\pm$0.2}               \\ \hline\hline
\multirow{5}{*}{ImageNet}      & FLOPS         & 260M                  & 320M                                 & 400M                                 & 450M                               & 500M                                 & 600M                                 \\ \cline{2-8} 
                               & US-Nets~\cite{yu2019universally} & 70.6$\pm$0.3          & 71.6$\pm$0.2                         & 71.8$\pm$0.1                         & 72.1$\pm$0.2                       & 72.3$\pm$0.4                         & 72.9$\pm$0.2                         \\ \cline{2-8} 
                               & Joslim~\cite{chin2021joslimrudydiana}  & 70.8$\pm$0.1          & 71.9$\pm$0.3                         & 72.5$\pm$0.2                         & 72.7$\pm$0.2                       & 72.9$\pm$0.2                         & 73.4$\pm$0.3                         \\ \cline{2-8} 
                               & DS-Net~\cite{li2021dynamic_dsnet}         & 70.6$\pm$0.2          & 72.1$\pm$0.1                         & 72.5$\pm$0.3                         & 72.6$\pm$0.1                       & 73.0$\pm$0.2                         & 73.3$\pm$0.2                         \\ \cline{2-8} 
                               & \textbf{TIPS}   & \textbf{71.8$\pm$0.4} & \textbf{73.2$\pm$0.3} & \textbf{73.6$\pm$0.3} & \textbf{74$\pm$0.2} & \textbf{74.3$\pm$0.3} & \textbf{74.7$\pm$0.1} \\ \hline
\end{tabular}
}\vspace{0mm}
\end{center}
\end{table}

\begin{table}[h]
\caption{Comparison of Top-1 test accuracy vs. FLOPS (in millions/Giga [M/G]) with SOTA training methods on ResNet-34. Best results are shown with bold fonts. Results are averaged over three runs. \vspace{0mm}}
\label{tab:app_rn34_flopsacc_std}
\begin{center}\scalebox{0.95}{

\begin{tabular}{|c|c||c|c|c|c|c|c|}
\hline
\multirow{5}{*}{CIFAR100}      & FLOPS         & 120M                  & 180M                  & 200M                   & 220M                  & 240M                   & 260M                  \\ \cline{2-8} 
                               & US-Nets~\cite{yu2019universally} & 63.1$\pm$0.2          & 63.9$\pm$0.1          & 64.4$\pm$0.3           & 64.8$\pm$0.2          & 65.0$\pm$0.1           & 65.4$\pm$0.1          \\ \cline{2-8} 
                               & Joslim~\cite{chin2021joslimrudydiana}  & 65.8$\pm$0.1          & 66.2$\pm$0.3          & 66.7$\pm$0.4           & 67.0$\pm$0.2          & 67.3$\pm$0.4           & 67.4$\pm$0.4          \\ \cline{2-8} 
                               & DS-Net~\cite{li2021dynamic_dsnet}         & 64.4$\pm$0.3          & 65.9$\pm$0.1          & 66.2$\pm$0.3           & 66.4$\pm$0.1          & 66.5$\pm$0.3           & 66.6$\pm$0.1          \\ \cline{2-8} 
                               & \textbf{TIPS}   & \textbf{67.3$\pm$0.1} & \textbf{67.4$\pm$0.2} & \textbf{67.8$\pm$0.2}  & \textbf{67.9$\pm$0.3} & \textbf{68.1$\pm$0.2}  & \textbf{68.2$\pm$0.3} \\ \hline\hline
\multirow{5}{*}{Tiny-ImageNet} & FLOPS         & 130M                  & 190M                  & 220M                   & 250M                  & 270M                   & 300M                  \\ \cline{2-8} 
                               & US-Nets~\cite{yu2019universally} & 42.9$\pm$0.1          & 43.2$\pm$0.2          & 44.3$\pm$0.3           & 44.7$\pm$0.3          & 44.9$\pm$0.2           & 45.2$\pm$0.3          \\ \cline{2-8} 
                               & Joslim~\cite{chin2021joslimrudydiana}  & \textbf{44.9$\pm$0.2} & \textbf{45.0$\pm$0.4} & 45.3$\pm$0.4           & 45.4$\pm$0.3          & 45.5$\pm$0.3           & 45.8$\pm$0.2          \\ \cline{2-8} 
                               & DS-Net~\cite{li2021dynamic_dsnet}         & 41.8$\pm$0.3          & 43.0$\pm$0.1          & 43.8$\pm$0.2           & 43.9$\pm$0.3          & 44.1$\pm$0.4           & 44.2$\pm$0.1          \\ \cline{2-8} 
                               & \textbf{TIPS}   & 44.1$\pm$0.4          & 44.6$\pm$0.1          & \textbf{45.4$\pm$0.2}  & \textbf{45.8$\pm$0.1} & \textbf{45.9$\pm$0.1}  & \textbf{46.0$\pm$0.2} \\ \hline\hline
\multirow{5}{*}{ImageNet}      & FLOPS         & 1.5G                  & 2.2G                  & 2.8G                   & 3.0G                  & 3.2G                   & 3.6G                  \\ \cline{2-8} 
                               & US-Nets~\cite{yu2019universally} & 67.8$\pm$0.4          & 69.2$\pm$0.1          & 69.7$\pm$0.2           & 70.1$\pm$0.1          & 70.2$\pm$0.3           & 70.5$\pm$0.2          \\ \cline{2-8} 
                               & Joslim~\cite{chin2021joslimrudydiana}  & 68.0$\pm$0.2          & \textbf{69.4$\pm$0.4} & 69.6$\pm$0.4           & 70.0$\pm$0.1          & 70.2$\pm$0.1           & 70.4$\pm$0.1          \\ \cline{2-8} 
                               & DS-Net~\cite{li2021dynamic_dsnet}         & 66.0$\pm$0.6          & 67.0$\pm$0.3          & 68.8$\pm$0.1           & 69.4$\pm$0.2          & 69.9$\pm$0.1           & 70.0$\pm$0.2          \\ \cline{2-8} 
                               & \textbf{TIPS}   & \textbf{68.4$\pm$0.5} & 69.3$\pm$0.2          & \textbf{70.8$\pm$0.2} & \textbf{71.1$\pm$0.1} & \textbf{71.4$\pm$0.2} & \textbf{71.9$\pm$0.2} \\ \hline
\end{tabular}
}\vspace{0mm}
\end{center}
\end{table}

\subsection{Additional Results for Table~\ref{tab:mbnv2_main_flopsacc} and Table~\ref{tab:rn34_main_flopsacc}}\label{app:std_dev}
We show the std. dev. values in Table~\ref{tab:app_mbn2_flopsacc_std} and Table~\ref{tab:app_rn34_flopsacc_std} for MobileNet-v2 and ResNet34, respectively.

\subsection{Details of TIPS}\label{app:detail_tips}
Given the supernet, we randomly sample 8 subnetworks and obtain the adjacency matrices $\bm{A}_k$ as described in Eq.~\ref{eq:adj_mat_cases}. As discussed in Section~\ref{sec:exp_ablation}, the ranking among multiple paths remains the same with varying value of $\lambda$. For simplicity, we set $\lambda$ to `1' for the inter-subnetwork coupling matrix $\widetilde{\bm{Z}_{i,j}}$ in Eq.~\ref{eq:InterLayer} to build the hyper-adjacency matrix $\hat{\bm{H}}(\lambda)$. Then we construct the DTMC as described in Eq.~\ref{eq:normalmatrix} and solve Eq.~\ref{eq:properyofdtmc} to obtain the stationary state distribution. Next, we exploit the TAS and TPS analysis to characterize the important operations as discussed in Section~\ref{sec:approach_nnpath} and Section~\ref{sec:TIPS}. 

\subsection{Observations of DTMC-based Analysis}
Based on our experiments on MLPs, MobileNet-v2, and ResNet34 (see Sec.~\ref{sec:experiment}), we can draw the following conclusions:
\begin{itemize}
    \item The TPS values and the important operations identified by our framework depend on the specific structure of a given supernet. Hence, we need to conduct the DTMC-based analysis individually for different supernets in order to have a meaningful understanding of operations importance.
    \item Empirically, we found that for inverted bottleneck-based MobileNet-v2 supernet and BasicBlock-based ResNet supernet, the first convolution layer was more important and more channels were sampled at those layers.
\end{itemize}

\subsection{Societal Impact of TIPS}
Our method does accelerate the convergence speed of the training process and thus reduces the total training costs. Indeed, as shown in Fig.~\ref{fig:converge_bcs}(b,c) in the main paper, to achieve the same training loss, our method requires far fewer training epochs compared to previous SOTA methods (Joslim~\cite{chin2021joslimrudydiana} and US-Nets~\cite{yu2019universally}). Hence, our method is clearly more environment-friendly than SOTA and implicitly addresses an important societal concern.

\subsection{Comparison with One-shot NAS}

We remark that our method focuses on the training methods for anytime inference in order to improve the test accuracy of anytime inference for neural networks instead of improving the accuracy of single networks; this is the key difference between anytime inference and neural architecture search (NAS). 
To demonstrate the benefits of our proposed training method, we compare our proposed TIPS with the training method of the one-shot NAS method Once-For-All (OFA)~\cite{cai2019onceforall}.

\begin{table}[h]
\caption{Comparison of Top-1 test accuracy vs. FLOPS (in millions/Giga [M/G]) with representative one-shot NAS method OFA on MobileNet-v2 under the same training setup. The best results are shown with bold fonts. Results are averaged over three runs. \vspace{0mm}}
\label{tab:app_oneshotnas}
\begin{center}\scalebox{1}{
\begin{tabular}{|l||l|l|l|l|l|l|}
\hline
\textbf{\#FLOPs} & \textbf{260M} & \textbf{320M} & \textbf{400M} & \textbf{450M} & \textbf{500M} & \textbf{600M} \\ \hline\hline
\textbf{OFA}     & 70.4          & 71.4          & 72.3          & 72.8          & 73.4          & 74            \\ \hline
\textbf{TIPS}    & \textbf{71.8} & \textbf{73.2} & \textbf{73.6} & \textbf{74}   & \textbf{74.3} & \textbf{74.7} \\ \hline
\end{tabular}
}\vspace{0mm}
\end{center}
\end{table}

To make an apples-to-apples comparison with OFA, we took the official training code for OFA and then trained our MobileNet-v2-based supernet on ImageNet under the same setup as ours (150 epochs, batchsize=512). 
As shown in Table~\ref{tab:app_oneshotnas}, our proposed TIPS achieves far better than OFA \#FLOPs-accuracy tradeoffs consistently; \textit{e.g.}, when the FLOPs budget is 320M, TIPS has a 1.8\% higher accuracy than OFA, which is a significant improvement on ImageNet.

{

\subsection{Overhead of Network Switch at Runtime}\label{app:overhead}

In our method, we store only the supernet and the configuration of each subnetwork (\textit{i.e.}, only the \#channels values for the layers in the supernet). This way, we do \textit{not} need to store and load the pretrained weights of different subnetworks separately. We provide the pseudo-code in Algorithm~\ref{alg:dynamic_infer} to better illustrate how we conduct the inference of AnytimeNNs. 

\begin{algorithm}[H]
   \caption{Pseudo code: Inference of AnytimeNNs}
   \label{alg:dynamic_infer}
{\small

\begin{algorithmic}[1]
\STATE {\bfseries Input:} Supernet checkpoint $G$, Pareto-optimal subnetworks' width configurations {$\Theta$}
\STATE {\bfseries Run:}
\STATE Load the supernet checkpoint $G$
\STATE Load subnetworks' width configurations {$\Theta$}
\WHILE{Running inference}
\STATE Index the suitable subnetwork configurations $\theta$ from {$\Theta$}
\FOR{ each layer $i$ in $G$}
\STATE Load $C_{IN_{i}}$ and $C_{OUT_{i}}$ from $\theta$
\STATE  Set \#input channels to $C_{IN_{i}}$ 
\STATE  Set \#output channels to $C_{OUT_{i}}$ 
\ENDFOR
\WHILE{hardware resources budget doesn't change}
\STATE Run inference
\ENDWHILE
\ENDWHILE
\end{algorithmic}}
\end{algorithm}

To quantitatively demonstrate the hardware efficiency of our method, we use MobileNet-v2 as the supernet then select twelve Pareto-optimal subnetworks under different FLOP budgets. We calculate the storage costs of these subnetworks. Specifically, storing these twelve subnetworks separately requires 117.8MB in total. In contrast, in our method, the storage cost of all these subnetwork configuration is quite negligible, \textit{i.e.}, requiring only 1.9 KB in total (6176$\times$ smaller) since it only requires storing layerwise width information for each subnetwork. Hence, our method is very hardware-efficient as it has much less overhead than storing all these subnetworks individually.

We also verify the hardware efficiency of our method as follows: As shown in Algorithm~\ref{alg:dynamic_infer}, we only need to load the checkpoint for the supernet $G$ once and the Pareto-optimal subnetwork configurations. To switch the subnetwork, we just select the suitable subnetwork configuration and reconfigure the width value of each layer  (see line 6-11 in Algorithm~\ref{alg:dynamic_infer}). For the same twelve Pareto-optimal subnetworks from MobileNet-v2, on a NVIDIA RTX3090 GPU with PyTorch Framework, we repeat the switching process 1000 times. We measure that reloading a new subnetwork checkpoint consumes 287ms, on average. In contrast, in our method, it takes only a negligible 0.037ms, on average, to switch the subnetwork (7756$\times$ faster than reloading the subnetworks checkpoint). 

}

\section{Illustration of Our Modeling Method and Sampling Process}\label{app:illustrate}
\begin{figure}[h]
    \centering
    
    \includegraphics[width=0.8\textwidth]{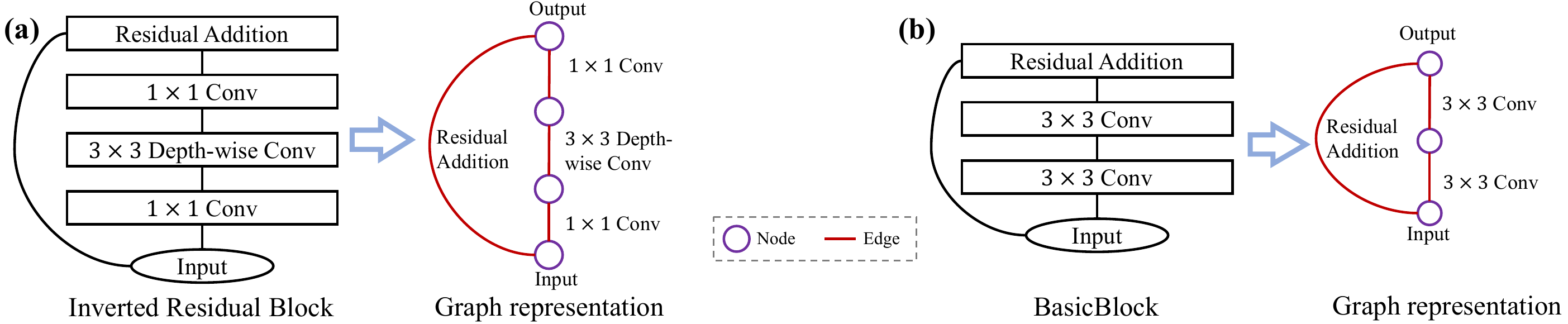}
    \caption{Illustration of how we model neural networks as graphs. \textbf{(a)} Inverted Residual block from MobileNet-v2~\cite{sandler2018mobilenetv2}. \textbf{(b)} BasicBlock from ResNet-18/34~\cite{veit2016residual}. As we mention in Section~\ref{sec:approach_model}, we model each operation (linear layers, convolutional layers, residual additions, pooling layers, etc.) as \textit{edges} in a graph; we model the input featuremaps and output featuremaps of these operations as \textit{nodes} in a graph.}
    \label{fig:rebuttal_network2graph}
\end{figure}

\begin{figure}[htb]
    \centering
    
    \includegraphics[width=0.8\textwidth]{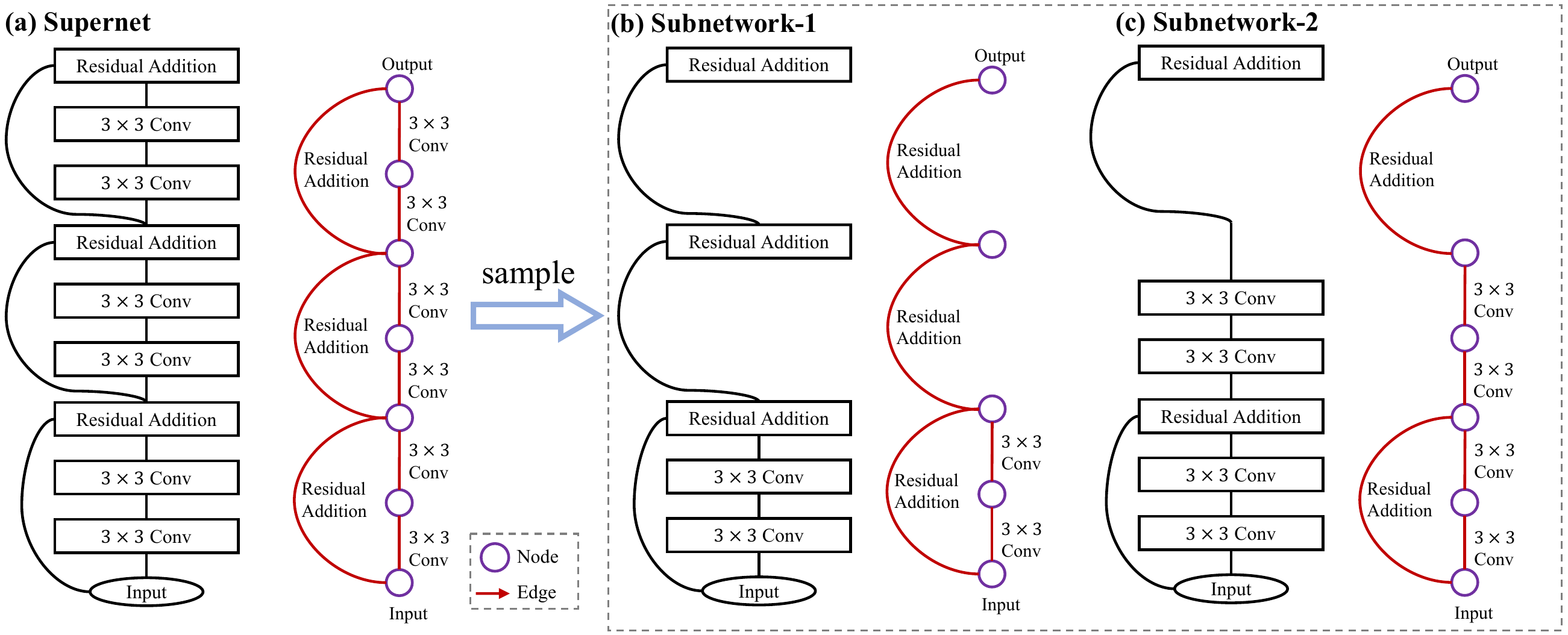}
    \caption{An illustration of sampling subnetworks and then converting subnetworks to graphs. We use a network with three BasicBlocks from ResNet18/34~\cite{veit2016residual}, for simplicity.}
    \label{fig:rebuttal_temporal}
\end{figure}

\subsection{Modeling Neural Networks as Graphs}
As shown in Fig,~\ref{fig:rebuttal_network2graph}, we illustrate how we model two commonly used blocks as graphs: Inverted Residual block from MobileNet-v2~\cite{sandler2018mobilenetv2} and BasicBlock from ResNet-18/34~\cite{he2016deepresnet}.

\subsection{Sampling Subnetworks from the Supernet}
As shown in Fig.~\ref{fig:rebuttal_temporal}, to further demonstrate how we model subnetworks as graphs, we use a network with three BasicBlocks as the supernet. Clearly, the same operation from the supernet can be skipped or kept in different subnetworks (this is temporally dependent). Our method captures these temporal relationships among multiple subnetworks; this is why we combine the adjacency matrices of multiple subnetworks into a hyper-adjacency matrix, as shown in Eq.~\ref{eq:superadj}. 

\begin{figure}[t]
    \centering
    
    \includegraphics[width=0.8\textwidth]{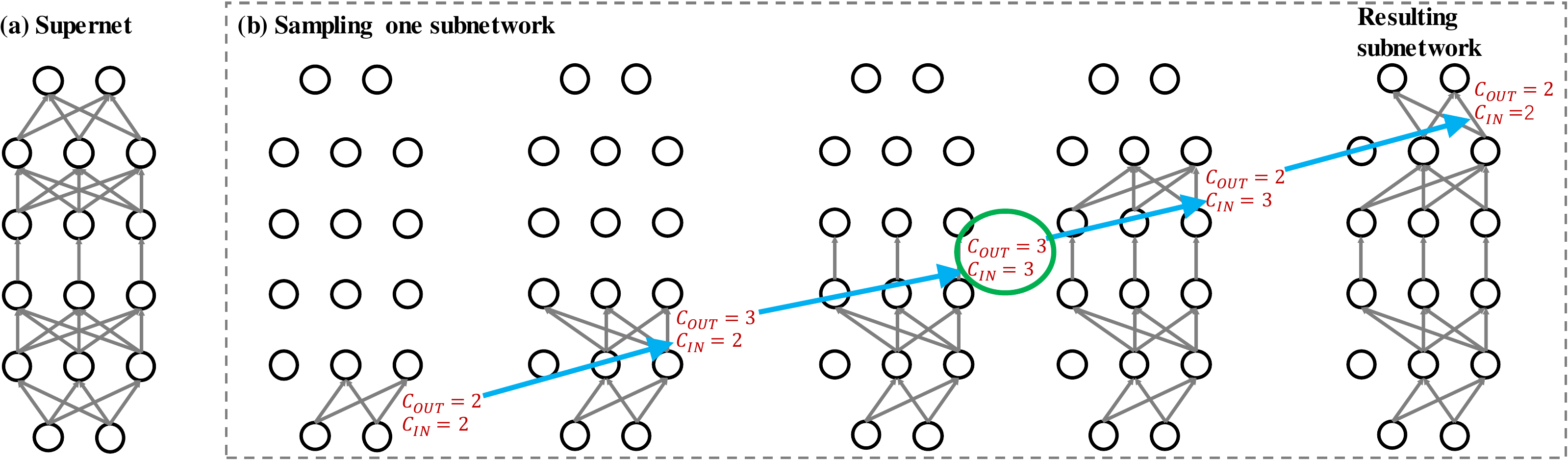}
    \caption{Sampling a subnetwork from the supernet. We show a supernet with 4 convolution layers and one depthwise convolution layer, for simplicity. We sample these layers based on input to output order in the supernet. The number of input channels of a given layer ($C_{IN}$) is always set to the same value as the number of output channels ($C_{OUT}$) of the previous layer; see the blue arrows in the figure. In particular, for a depthwise convolution layer, $C_{OUT}$ is always set to the values its $C_{IN}$; see the green circle in the figure.}
    \label{fig:rebuttal_sampling}
\end{figure}

\subsection{Illustration of Validity of Subnetworks}
As shown in Fig.~\ref{fig:rebuttal_sampling}, given the supernet, to ensure the validity of the sampled networks, we sample \#channels of each layer from input to output as follows:
\begin{enumerate}
    \item The number of input channels of a given layer is always set to the same value as the number of output channels of the previous layer (see the blue arrows in Fig.~\ref{fig:rebuttal_sampling}). 
    \item For a depthwise convolution layer, we always set the number of output channels to the same value as its input channels (see the green circle in Fig.~\ref{fig:rebuttal_sampling}). 
\end{enumerate}

\end{document}